\documentclass[letterpaper]{article} 
\usepackage{aaai2027}  
\usepackage[hyphens]{url}  
\usepackage{graphicx} 
\usepackage{natbib}  
\usepackage{caption} 
\usepackage{algorithm}
\usepackage{algorithmic}
\usepackage{graphicx}
\usepackage{makecell}
\usepackage{amsmath}
\usepackage{booktabs} 
\usepackage{multirow}
\usepackage{amssymb}
\usepackage{newfloat}
\usepackage{listings}
\DeclareCaptionStyle{ruled}{labelfont=normalfont,labelsep=colon,strut=off} 
\floatstyle{ruled}
\newfloat{listing}{tb}{lst}{}
\floatname{listing}{Listing}

\usepackage{booktabs}

\title{LAWM-3D: Learning 3D-Aware Latent Actions from Human Videos for Generalizable Robot World Models}

\author {
Jiarui Yang\textsuperscript{\rm 1,\rm 2}, Jiale Zhange\textsuperscript{\rm 2}, Jiawei Li\textsuperscript{\rm 2}, Hang Guo\textsuperscript{\rm 4}, Wen Huang\textsuperscript{\rm 2}, Jinpeng Wang\textsuperscript{\rm 3}, Peidong Liu\textsuperscript{\rm 2}, Shu-Tao Xia\textsuperscript{\rm 2}
}
\affiliations {
    \textsuperscript{\rm 1} Nankai University 
    \textsuperscript{\rm 2} Tsinghua University
    \textsuperscript{\rm 3} Harbin Institute of Technology \\
    \textsuperscript{\rm 4} Swiss Federal Institute of Technology in Lausanne (EPFL) 
}

\begin{document}

\maketitle

\begin{figure*}[!ht]
  \centering
  \includegraphics[width=0.8\textwidth]{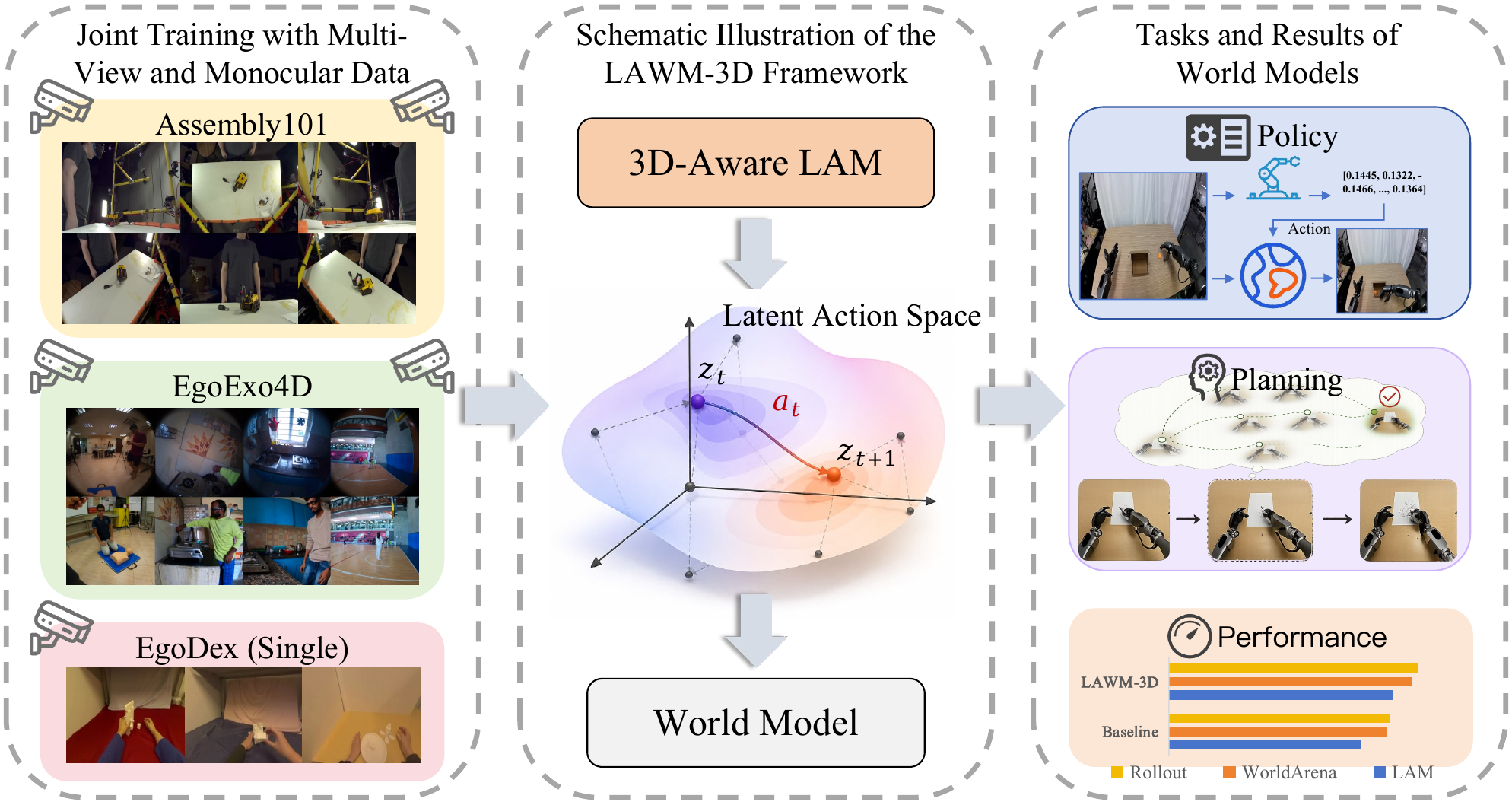}
  \caption{
  \textbf{Overview of LAWM-3D framework.}
  The model learns 3D-aware latent actions through joint training on multi-view and single-view datasets. The learned latent action space enables world model pretraining and supports downstream tasks, achieving superior overall performance compared to the baseline method DreamDojo \citep{dreamdojo}.
  }
  \label{fig:teaser}
\end{figure*}

\begin{abstract}


World models enable agents to perform forward rollout and planning without real-world interaction. However, their application in open-world embodied intelligence remains limited by the high cost of action annotations and the heterogeneity of action spaces across platforms. Recently, latent action models (LAMs) have alleviated this bottleneck by learning action representations directly from unlabeled human videos in a self-supervised manner. Nevertheless, most existing LAMs rely on single-view inputs and operate primarily in 2D pixel space, raising a fundamental question: can simply incorporating multi-view videos into LAM training endow the learned latent actions with 3D-aware perception? Our study shows that the answer is negative. The primary reasons lie in future-frame appearance leakage as well as inter-camera appearance discrepancies and viewpoint variations. To address these issues, we propose LAWM-3D, which introduces three tightly coupled key designs: (1) a multi-view invariant unified action tokenization scheme for learning 3D-aware latent actions; (2) a geometric alignment constraint that anchors intermediate encoder features to a pretrained 3D foundation model, thereby explicitly providing cross-view geometric correspondences; and (3) a non-injective RGB–D joint reconstruction objective that prevents shortcut learning from future-frame appearance information, forcing the LAM to focus supervision on motion cues with geometric significance. Importantly, these components are not simply stacked but are tightly coupled through a unified motivation. Built upon a two-stage paradigm of large-scale human video pretraining followed by robot fine-tuning, extensive experiments demonstrate that the proposed 3D-aware latent actions significantly improve world model performance, achieving SOTA results in generation quality, physical consistency, and generalization ability.

\end{abstract}



\section{Introduction}





World models serve as internal simulators of environmental dynamics, enabling agents to perform action-conditioned rollouts without interacting with the real world, thereby supporting planning and decision-making \citep{worldmodel}. While this paradigm has shown strong potential in data-rich closed environments, its generalization in open-world embodied intelligence remains an open challenge. Existing works typically rely on large-scale datasets with explicit action annotations, which are expensive to collect and difficult to scale \citep{ding2025understanding}. In addition, substantial heterogeneity across robotic platforms and limited diversity in current datasets further hinder effective transfer to novel environments and tasks \citep{min2024driveworld}.

Recently, latent action models (LAMs) have been proposed to alleviate these limitations by learning action representations from large-scale unlabeled human videos \citep{adaworld, lawm, lapv}. For example, DreamDojo \citep{dreamdojo} formulates LAMs within a variational autoencoder (VAE) framework, where latent actions are inferred from consecutive frames to reconstruct future observations, encouraging the latent space to capture key state transitions. However, most existing methods operate on single-view videos \citep{egodex} and primarily model 2D appearance changes, lacking explicit reasoning about 3D motion and interactions. As a result, the learned latent actions often fail to reflect true 3D motion semantics, limiting downstream world model performance in prediction and planning tasks \citep{guerry2017snapnet}.

While incorporating multi-view data is a natural way to improve 3D understanding, we find that naive integration is ineffective. First, performing latent action extraction in pixel space introduces substantial task-irrelevant information, leading the model to prioritize pixel reconstruction over motion semantics \citep{saha2025pixels}. Second, because LAMs are trained with adjacent-frame objectives, future-frame RGB information can be directly exploited by the encoder, causing latent actions to degenerate into compressed representations of appearance rather than motion. Finally, multi-view observations are 2D projections without explicit cross-view geometric constraints, making consistent 3D representation learning difficult \citep{chen2025learning}.

To tackle these challenges, we propose LAWM-3D, as shown in Fig. \ref{fig:teaser}. We introduce a unified action tokenization scheme and jointly train on multi-view and single-view data to learn view-invariant latent actions. We further propose a non-injective RGB–D joint prediction objective, which mitigates future-frame RGB leakage and reduces reliance on view-dependent appearance cues, encouraging the model to focus on geometrically meaningful motion. In addition, we incorporate a pretrained 3D foundation model and enforce geometric alignment constraints to explicitly model cross-view correspondences. 

Building upon these designs, we pretrain world models on large-scale human videos using the learned latent actions and further adapt them to robotic tasks via post-training for action space alignment. Extensive qualitative and quantitative experiments show that LAWM-3D consistently outperforms prior LAM methods. In particular, evaluations across 16 benchmark metrics and out-of-distribution settings demonstrate significant improvements in generation quality, physical consistency, action controllability, and 3D spatial understanding, as well as stronger cross-scene generalization and interaction modeling ability.

\begin{figure*}[t]
\centering
\includegraphics[width=2\columnwidth]{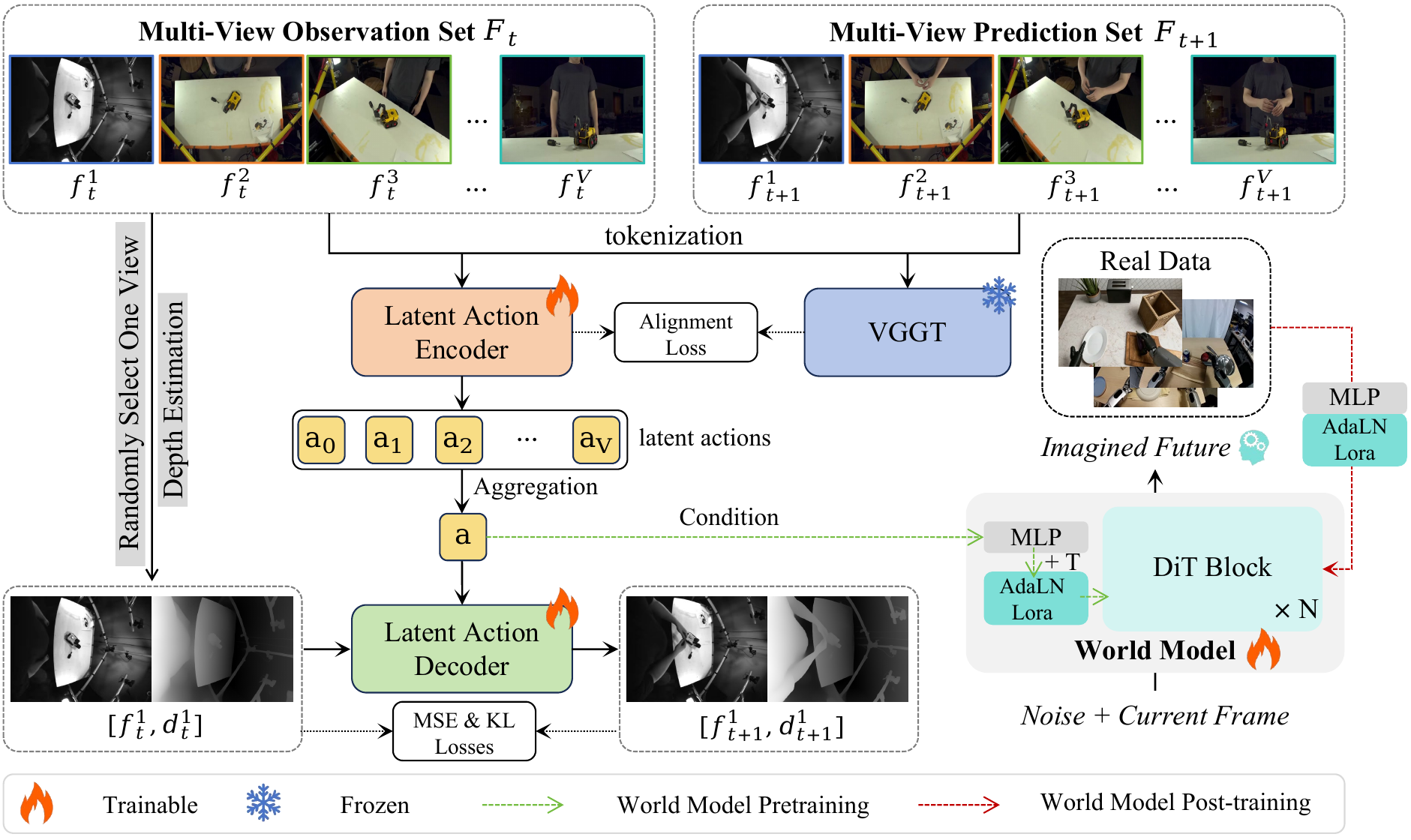}
\caption{Architecture of the 3D-aware latent action model and world model training pipeline.}
\label{fig: fw}
\end{figure*}

\section{Related Work}

\textbf{Latent Action Models.} LAMs aim to learn implicit action representations from unlabeled videos and have recently attracted significant attention in world models and embodied intelligence. Genie \citep{bruce2024genie} model latent actions as conditional control signals for controllable video generation and environment dynamics prediction, while approaches like CoMo \citep{yang2025learning} and DreamDojo \citep{dreamdojo} treat them as unified intermediate action representations to reduce reliance on expensive action annotations and improve cross-task generalization. Despite the demonstrated effectiveness of this paradigm, most existing methods still lack explicit modeling of 3D spatial structure and motion geometry. To address this limitation, recent studies have introduced geometric awareness. For example, UniLACT \citep{govind2026unilact} incorporates depth supervision under a single-view setting to enhance 3D understanding, while MVP-LAM \citep{lee2026mvp} leverages cross-view reconstruction constraints to learn view-invariant action representations. However, unlike UniLACT, we employ non-injective depth supervision, which strengthens geometric constraints while avoiding direct information injection, thereby reducing shortcut learning. In contrast to MVP-LAM, which relies on synthetic or strictly aligned multi-view settings, we model real-world human multi-view videos and further introduce a unified action tokenization and coupled learning mechanism to enforce cross-view consistency, making our approach more suitable for real-world scenarios with strong viewpoint heterogeneity.


\textbf{3D-Aware World Models.} Recent research has increasingly shifted toward developing 3D-aware world models to improve multi-view consistency and geometric plausibility in generated results. A line of work enhances spatial understanding through geometric constraints and 3D representations. For instance, Matrix-3D \citep{yang2025matrix} and WorldStereo \citep{zhang2026worldstereo} alleviate cross-view inconsistencies via panoramic representations, geometric routing mechanisms, and geometric memory modules, respectively. Building upon this direction, Geometry Forcing \citep{wu2026geometry} further improves geometric consistency by aligning video diffusion features with pretrained 3D foundation models, while FantasyWorld \citep{dai2025fantasyworld} introduces a trainable geometric branch on top of a frozen video foundation model to jointly model video latents and implicit 3D fields. Beyond static scene modeling, recent efforts have also begun to investigate the temporal modeling of dynamic 3D environments. For example, TesserAct \citep{zhen2025tesseract} jointly generates RGB, depth, and normal sequences to predict the temporal evolution of 3D scenes under action conditioning, whereas Kinema4D \citep{xu2026kinema4d} combines RGB observations with robotic point cloud sequences to model the dynamics of evolving point clouds.

\section{Method}
\subsection{Overview}

As illustrated in Fig. \ref{fig: fw}, \textbf{LAWM-3D} consists of a latent action autoencoder \citep{kingma2013auto} and a conditional world model, and is trained in a staged manner to improve generalization and cross-domain adaptability. The overall training pipeline comprises three stages:
(1) training a LAM on large-scale multi-view and monocular video data to extract view-invariant action representations;
(2) pretraining the world model on human videos with corresponding pseudo-action labels to learn the dynamics mapping from actions to future observations;
(3) fine-tuning on real robot interaction data to align latent actions with actual control signals, enabling downstream control and planning tasks.

\subsection{3D-Aware Latent Action Model}

We formulate the LAM as a multi-view variational autoencoder to learn view-invariant latent action representations from multi-view observations, enabling consistent action semantics in 3D space. Given a set of synchronized observations from multiple camera views,
\begin{equation}
F_t = \{ f_t^v \}_{v=1}^{V},
\end{equation}
which includes one ego view and $V$ exo views, the model takes consecutive observations $F_{t:t+1}$ as input and learns a latent action $\textbf{a}$ that captures the dynamics from time $t$ to $t+1$.

Specifically, input frames are first divided into $16 \times 16$ patches and mapped into a sequence of tokens. These tokens are concatenated with action tokens $\textbf{a}_{t:t+1}$, and augmented with view embeddings and rotary positional encodings \citep{heo2024rotary}. The view embeddings distinguish different camera viewpoints, while temporal information is implicitly modeled via positional encoding. 

The encoder adopts a spatio-temporal Transformer architecture \citep{bruce2024genie} with $L$ layers, where each layer alternates between spatial self-attention and temporal self-attention, followed by feed-forward networks. Due to the bidirectional nature of attention, representations at time $t+1$ can implicitly incorporate information from time $t$. Therefore, for frame pairs $f_{t:t+1}$ under the same view, we retain only the encoded $a_{t+1}$ token. These tokens are then aggregated across views (e.g., via pooling) to obtain a unified latent representation, which is mapped to Gaussian parameters $(\mu, \log \sigma)$. The final latent action $\textbf{a}$ is obtained via the reparameterization trick. The decoder is implemented as a spatial Transformer, which predicts the future frame $f_{t+1}$ conditioned on the current frame $f_t$ under view $v$ and the latent action $\textbf{a}$. The overall training objective follows the $\beta$-VAE formulation \citep{dreamdojo}:
\begin{equation}
\begin{aligned}
\mathcal{L}_{\text{Rec}}(f^v_{t+1})
&=
\mathbb{E}_{q_{\phi}(\hat{\textbf{a}}\mid f_{t:t+1}^v)}
\left[
\log p_{\theta}(f_{t+1}^v\mid \textbf{a}, f^v_{t})
\right] \\
&\quad
- \beta \, D_{\mathrm{KL}}
\left(
q_{\phi}(\textbf{a}\mid f^v_{t:t+1}) \,\|\, p(\textbf{a})
\right).
\end{aligned}
\end{equation}

\subsubsection{Joint Training with Single and Multi-View Data}

We posit that actions in 3D space exhibit invariance across viewpoints. Accordingly, the model should capture motion structures that are consistent across views, rather than relying on view-specific 2D pixel variations. To this end, during training, we apply a random view masking strategy to the input view set $F_{t:t+1}$, where at least one view is retained. In the decoding stage, a view is randomly selected from the retained subset for reconstruction. This design forces the encoder to infer a consistent latent action representation under incomplete observations, thereby enhancing robustness to viewpoint variations. Benefiting from the shared latent action representation, LAM can be trained jointly on a mixture of multi-view and monocular data distributions. At inference time, the model naturally degrades to single-view input while maintaining stable performance.

\begin{figure}[t]
\centering
\includegraphics[width=1\columnwidth]{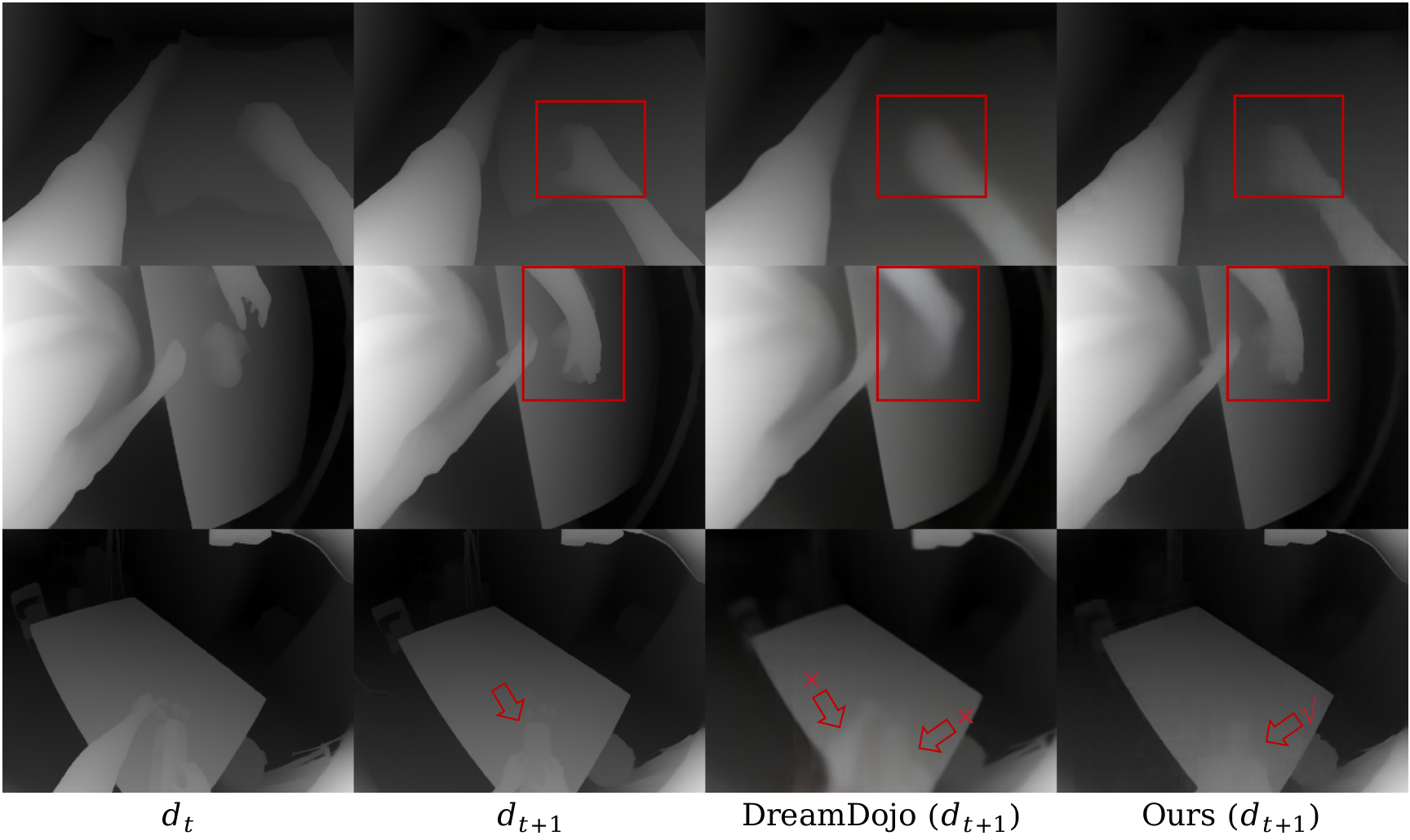}
\caption{Visual comparison of depth prediction on ego-view. Our method demonstrates stronger depth understanding. The third row highlights the accuracy of large-motion prediction.}
\label{fig: lam}
\end{figure}

\begin{table}[t]
\centering
\caption{Quantitative comparison of latent action models on Ego-Exo4D and Assembly101 datasets.}
\label{tab:rgb_depth_compare}

\renewcommand{\arraystretch}{1.08}
\setlength{\tabcolsep}{3.2pt}

\resizebox{\columnwidth}{!}{
\begin{tabular}{c|cc|cc|cc|cc}
\toprule

\multirow{3}{*}{Method}

& \multicolumn{4}{c|}{EgoExo4D}
& \multicolumn{4}{c}{Assembly101} \\

\cmidrule(lr){2-5}
\cmidrule(lr){6-9}

& \multicolumn{2}{c|}{RGB}
& \multicolumn{2}{c|}{Depth}
& \multicolumn{2}{c|}{RGB}
& \multicolumn{2}{c}{Depth} \\

\cmidrule(lr){2-3}
\cmidrule(lr){4-5}
\cmidrule(lr){6-7}
\cmidrule(lr){8-9}

& PSNR$\uparrow$
& SSIM$\uparrow$
& PSNR$\uparrow$
& SSIM$\uparrow$
& PSNR$\uparrow$
& SSIM$\uparrow$
& PSNR$\uparrow$
& SSIM$\uparrow$ \\

\midrule

LAPA
& 30.75 & 0.842
& / & /
& 31.26 & 0.867
& / & / \\

UniLACT
& 30.97 & 0.847
& 32.58 & 0.880
& 31.66 & 0.875
& 33.91 & 0.911 \\

CoMo
& 31.85 & 0.864
& / & /
& 32.96 & 0.891
& / & / \\

DreamDojo
& 32.91 & 0.901
& / & /
& 33.43 & 0.922
& / & / \\

DreamDojo-D
& 32.56 & 0.893
& 33.54 & 0.907
& 33.12 & 0.907
& 34.35 & 0.921\\

LAWM-3D
& \textbf{33.38} & \textbf{0.936}
& \textbf{35.62} & \textbf{0.939}
& \textbf{35.76} & \textbf{0.944}
& \textbf{35.85} & \textbf{0.948} \\

\bottomrule
\end{tabular}
}

\end{table}

\subsubsection{Pixel-Agnostic Learning via Depth Map}


Although the latent action $\textbf{a}$ is designed to be view-invariant, relying solely on pixel reconstruction loss introduces a critical limitation. In practice, the model can be biased by view-dependent factors such as appearance variations, occlusions, and imaging conditions, leading it to overfit low-level visual statistics. More importantly, since the encoder takes frame pairs $f_{t:t+1}$ as input, a pure reconstruction objective may encourage shortcut learning, where RGB information from the future frame $f_{t+1}$ is directly compressed into the latent variable $\textbf{a}$. In this case, $\textbf{a}$ degenerates into an encoding of future pixel distributions rather than a representation of action semantics, thereby undermining the predictive capability of the world model.

To address this issue, we introduce non-injective depth prediction as an auxiliary supervision signal in the decoding stage, which is not accessible to the encoder. Specifically, the decoder takes as input the concatenation of the current frame and its depth map, and jointly predicts $\{f_{t+1}, d_{t+1}\}$. The model is optimized using a joint reconstruction loss $\mathcal{L}_{\text{Rec}}(\{f^v_{t+1}, d^v_{t+1}\})$. Depth provides view-consistent geometric cues and is less sensitive to appearance variations. This mechanism effectively suppresses RGB information leakage and encourages the model to focus on geometrically meaningful dynamic regions, thereby improving the semantic quality of the learned latent action representation.

\subsubsection{Geometric Representation Alignment}



Reconstruction and depth supervision alone are insufficient to fundamentally enforce the learning of cross-view consistent 3D structures. To this end, we further introduce a geometry alignment constraint to explicitly inject 3D priors. Specifically, inspired by Geometry Forcing \citep{wu2026geometry}, we leverage a pretrained 3D foundation model (VGGT \citep{wang2025vggt}) to align the geometric features of the encoder. Unlike Geometry Forcing that aligns all VGGT features to diffusion models \citep{ho2020denoising}, we focus on aligning the intermediate spatial-semantic features of a ViT encoder, which more directly constrain geometric consistency. Let the intermediate features of the encoder be $h \in \mathbb{R}^{K \times V \times P \times D}$, where $K$ is the number of layers, $V$ is the number of views, $P$ is the number of patches, and $D$ is the feature dimension. The corresponding geometric features from VGGT are denoted as $
y \in \mathbb{R}^{K \times V \times P' \times D'}$.

Since VGGT explicitly models multi-view relationships via alternating intra-frame and inter-frame attention, its features inherently encode cross-view consistent 3D structures. We first map $h$ into the same feature space as $y$ via a projection head $f$, and apply a cosine similarity loss for directional alignment:

\begin{equation}
    \mathcal{L}_{\text{Angular}} 
= - \frac{1}{K V P} \sum_{k,v,p} 
\cos \big( y_{k,v,p}, \, f_{\phi}(h_{k,v,p}) \big).
\end{equation}

Considering that directional alignment alone may ignore magnitude information, while directly applying MSE can lead to training instability \citep{wu2026geometry}, we further introduce a prediction head $g$ to regress the full geometric representation on normalized features:

\begin{equation}
    \mathcal{L}_{\text{Scale}} 
= \frac{1}{K V P} \sum_{k,v,p} 
\left\| 
g_{\psi} \left( 
\frac{f_{\phi}(h_{k,v,p})}{\|f_{\phi}(h_{k,v,p})\|_2} 
\right) 
- y_{k,v,p} 
\right\|_2^2.
\end{equation}

Finally, the overall loss consists of the joint reconstruction loss and weighted alignment losses:

\begin{equation}
\mathcal{L_{\text{LAM}}} = \mathcal{L}_{\text{Rec}} + \lambda_{\text{Angular}} \mathcal{L}_{\text{Angular}} + \lambda_{\text{Scale}} \mathcal{L}_{\text{Scale}}.
\end{equation}

\begin{table*}[t]
\centering
\caption{Comprehensive video evaluation results across 16 metrics spanning six dimensions. Best results are \textbf{bolded}.}
\label{tab:arena}
\resizebox{\textwidth}{!}{%
\begin{tabular}{l|ccc|ccc|ccc|cc|cc|ccc}
\toprule
\multirow{3}{*}{Methods}
  & \multicolumn{3}{c|}{\textbf{Visual Quality}}
  & \multicolumn{3}{c|}{\textbf{Motion Quality}}
  & \multicolumn{3}{c|}{\textbf{Content Consistency}}
  & \multicolumn{2}{c|}{\textbf{Physics Adherence}}
  & \multicolumn{2}{c|}{\textbf{3D Accuracy}}
  & \multicolumn{3}{c}{\textbf{Controllability}} \\
\cmidrule(lr){2-4}\cmidrule(lr){5-7}\cmidrule(lr){8-10}
\cmidrule(lr){11-12}\cmidrule(lr){13-14}\cmidrule(lr){15-17}
  & \makecell{Image\\Quality}
  & \makecell{Aesthetic\\Quality}
  & \makecell{JEPA\\Similarity}
  & \makecell{Dynamic\\Degree}
  & \makecell{Flow\\Score}
  & \makecell{Motion\\Smoothness}
  & \makecell{Subject\\Consistency}
  & \makecell{Background\\Consistency}
  & \makecell{Photometric\\Consistency}
  & \makecell{Interaction\\Quality}
  & \makecell{Trajectory\\Acc.}
  & \makecell{Depth\\Acc.}
  & Perspectivity
  & \makecell{Instruction\\Follow.}
  & \makecell{Semantic\\Align.}
  & \makecell{Action\\Follow.} \\
\midrule
GigaWorld
  & 0.463 & 0.392 & 0.437
  & 0.614 & 0.315 & 0.778
  & 0.734 & 0.832 & 0.179
  & 0.529 & 0.159
  & 0.618 & 0.760
  & 0.607 & 0.851 & \textbf{0.116} \\
Genie
  & 0.229 & 0.326 & 0.334
  & \textbf{0.674} & 0.090 & 0.694
  & 0.770 & 0.883 & 0.205
  & 0.205 & 0.074
  & 0.851 & 0.520
  & 0.208 & 0.855 & 0.018 \\
RoboMaster
  & 0.353 & 0.405 & 0.302
  & 0.658 & 0.153 & 0.700
  & 0.810 & 0.903 & 0.341
  & 0.533 & 0.122
  & 0.809 & 0.738
  & 0.569 & 0.846 & 0.082 \\
Cosmos 2.5
  & 0.440 & 0.355 & 0.903
  & 0.592 & 0.260 & 0.735
  & 0.804 & 0.895 & 0.355
  & 0.541 & 0.294
  & 0.862 & 0.763
  & 0.611 & 0.857 & 0.070 \\
DreamDojo
  & 0.476 & \textbf{0.411} & 0.915
  & 0.603 & 0.304 & 0.771
  & 0.812 & 0.909 & 0.356
  & 0.558 & 0.322
  & 0.873 & 0.795
  & 0.680 & 0.883 & 0.087 \\
WoW
  & 0.460 & 0.381 & 0.744
  & 0.454 & 0.271 & 0.781
  & 0.812 & 0.893 & 0.221
  & 0.538 & 0.212
  & 0.727 & 0.749
  & 0.573 & 0.881 & 0.049 \\
IRASim
  & 0.347 & 0.365 & 0.916
  & 0.419 & 0.211 & 0.692
  & 0.813 & 0.898 & 0.350
  & 0.567 & 0.265
  & 0.870 & 0.782
  & 0.648 & 0.859 & 0.059 \\
\midrule
\textbf{Ours}
  & \textbf{0.491} & 0.397 & \textbf{0.920}
  & 0.618 & \textbf{0.320} & \textbf{0.789}
  & 0.815 & \textbf{0.913} & \textbf{0.361}
  & \textbf{0.578} & \textbf{0.350}
  & \textbf{0.889} & \textbf{0.805}
  & \textbf{0.731} & \textbf{0.892} & 0.095 \\
\bottomrule
\end{tabular}%
}
\end{table*}

\subsection{World Model}


Latent actions with 3D-aware representations better capture the dynamics of the real world, thereby improving the effectiveness of world model pretraining. To address the generalization limitations caused by the scarcity of real robotic data, we leverage human videos and pseudo-actions for pretraining. We adopt Cosmos-Predict2.5 \citep{ali2025world} as the backbone world model. During pretraining, human videos are first segmented into temporal chunks, from which latent action representations are extracted and concatenated into action sequences. These latent action chunks are then projected via a multilayer perceptron (MLP) into the same dimensional space as the timestep embeddings of the world model. The projected action embeddings are subsequently added to the timestep embeddings and jointly injected into the adaptive layer normalization (AdaLN) modules in each DiT block \citep{peebles2023scalable}.

During the post-training stage, we use the initial pose of each latent video frame as a reference anchor. The action space is reformulated into a relative action space by computing pose increments with respect to this reference, and the same injection mechanism is applied. Notably, we reinitialize the first layer of the action MLP and fully fine-tune it jointly with all other pretrained parameters. In addition to the standard Cosmos training objective $\mathcal{L}_{\text{flow}}$, we further incorporate the temporal consistency loss proposed in DreamDojo to enhance temporal coherence. The overall training objective is therefore defined as follows:

\begin{equation}
\small
\mathcal{L}_{\text{WM}}(\theta) = \mathcal{L}_{\text{flow}}(\theta) +  \lambda_{wm}\mathbb{E}\left[\sum_{i=1}^{K-1} \left\|(\hat{v}^{i+1} - \hat{v}^{i}) - (v^{i+1} - v^{i})\right\|^2\right].    
\end{equation}
where $v^i$ is the velocity of the $i$-th frame within the video latent sequence.

\begin{figure}[t]
\centering
\includegraphics[width=1\columnwidth]{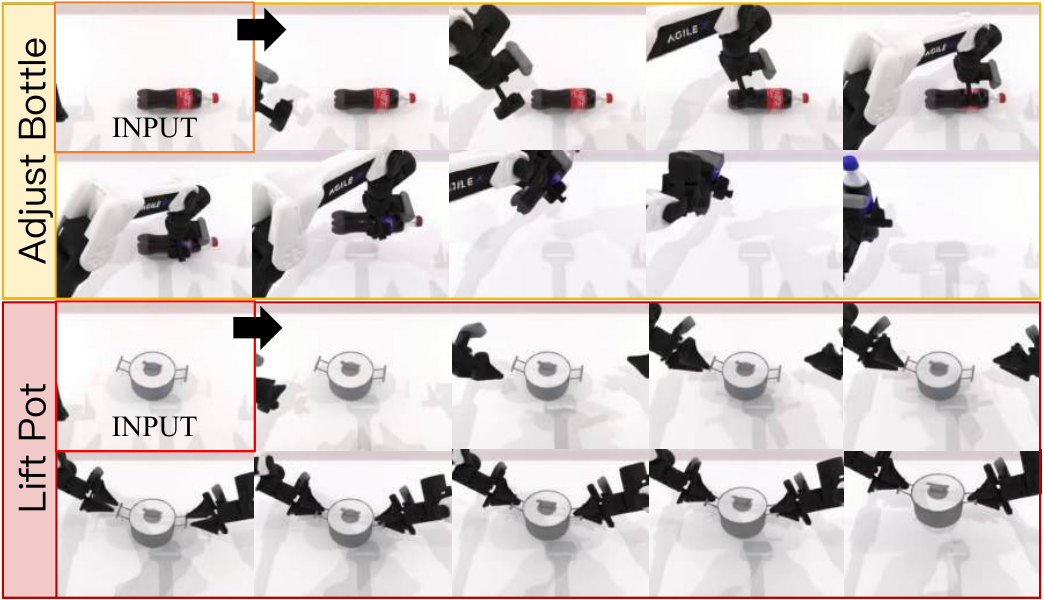}
\caption{Qualitative visualization of world model rollouts on robotic manipulation tasks in the WorldArena benchmark.}
\label{fig: arena}
\end{figure}

\begin{figure*}[t]
\centering
\includegraphics[width=2\columnwidth]{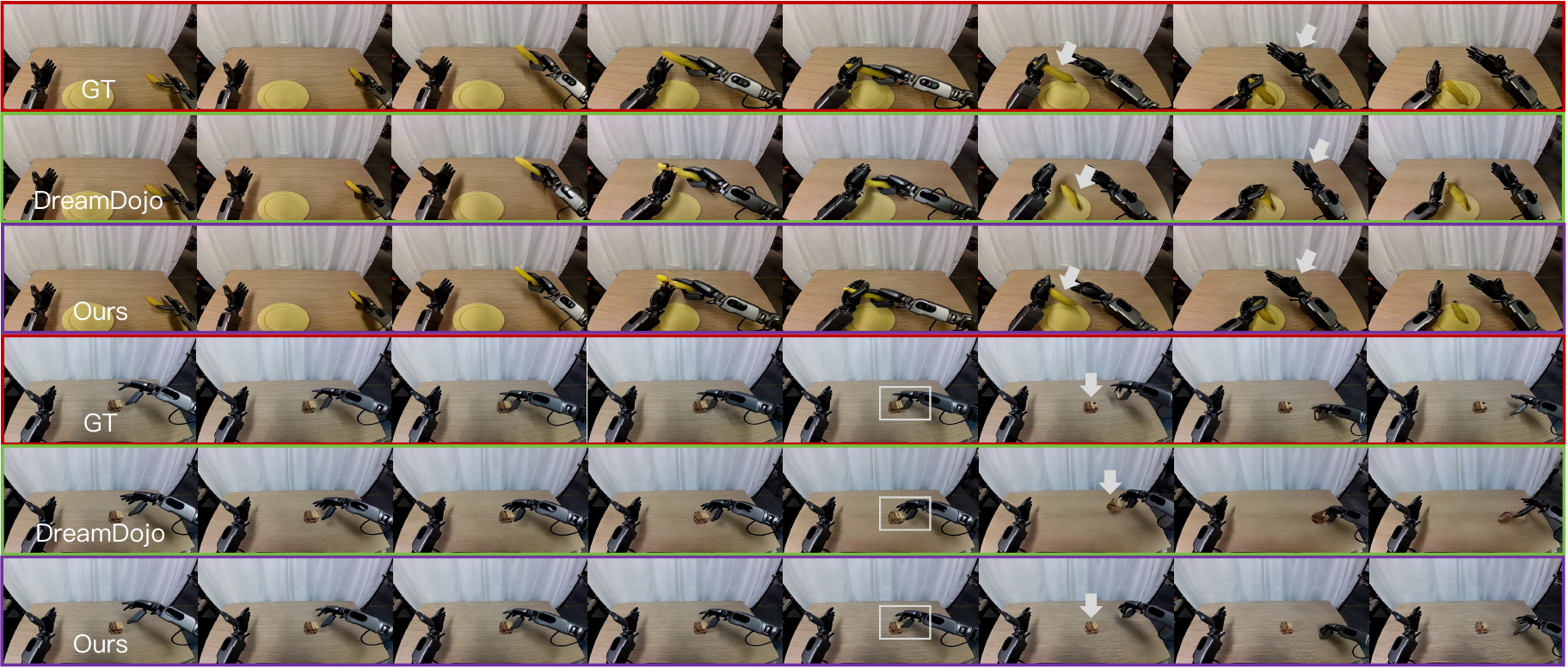}
\caption{Comparison of real-world rollouts between ground truth (GT), DreamDojo, and our method on manipulation scenarios. Our method exhibits stronger physical plausibility in object interactions (e.g., interactions with the banana and in the spatial positioning of blocks), demonstrating that 3D-aware latent actions help world models learn real-world physics and 3D spatial structure more effectively.}
\label{fig: ood}
\end{figure*}

\section{Experiments}





\subsection{Implementation Details}
Both the encoder and decoder consist of $L=24$ layers. We align the intermediate representations extracted from the encoder and VGGT at layers $h=[6,11]$. The latent action representation $\textbf{a}$ is set to a dimensionality of 32. We set $\beta=10^{-6}$, while the weights for the angular and scale losses are empirically set to 0.5 and 0.05, respectively. For LAM training, we employ AdamW \citep{loshchilov2017decoupled} with a weight decay of 0.01 and a constant learning rate of $2.5 \times 10^{-5}$. For world model training, we set $\lambda_{wm}=0.1$ and adopt AdamW, with a weight decay of 0.1 and a learning rate of $1.6 \times 10^{-4}$. An exponential moving average (EMA) of model parameters is maintained throughout training and is used for all evaluations and result generation. Both the LAM and the world model are pretrained from scratch on H20 GPUs. Specifically, the LAM is trained with a batch size of 128 for 600k iterations, while the world model is trained with a batch size of 256 for 300k iterations.

\vspace{-0.2cm}
\subsection{Training Datasets}
We train our model on multiple human manipulation video datasets, including the multi-view datasets Ego-Exo4D \citep{egoexo4d} and Assembly101 \citep{assembly101}, as well as the single-view dataset EgoDex \citep{egodex}. Ego-Exo4D contains approximately 1,286 hours of video collected from 740 participants across 123 different environments in 13 cities worldwide. Each sequence simultaneously captures egocentric views and 4--5 synchronized third-person views, covering a wide range of skill-intensive activities such as cooking, sports, and musical performance. Assembly101 is a multi-view procedural activity dataset consisting of 4,321 toy assembly and disassembly videos with a total duration of approximately 513 hours, captured using 8 static cameras and 4 head-mounted egocentric cameras. EgoDex is a single-view egocentric dexterous manipulation dataset containing 829 hours of video, approximately 90 million frames, and 338k task demonstrations, covering 194 tabletop manipulation tasks.

We apply a unified random split strategy to all datasets, partitioning the training and testsets with a ratio of 10:1. The pretraining of the latent action model additionally incorporates AgiBot-World \citep{bu2025agibot}, RT-1 \citep{brohan2022rt}, Language-Table \citep{nair2022learning}, and DROID \citep{khazatsky2024droid}. The maximum number of views is set to $V=7$, and the input video resolution is fixed at $320 \times 240$. To improve the model's ability to capture actions across different temporal scales, we randomly apply temporal downsampling with factors ranging from $1\sim4\times$ to construct training samples with varying motion intervals. For world model pretraining, we uniformly use single-view video inputs with a resolution of $640 \times 480$. All videos are processed into sequences of 13 frames, where the first frame is used as the conditioning input and the remaining frames are treated as future targets for prediction.

\subsection{Comparison of LAMs}


We compare our method with recent SOTA LAMs, including LAPA \citep{lapa}, UniLACT \citep{govind2026unilact}, CoMo \citep{yang2025learning}, MVP-LAM \citep{lee2026mvp}, and DreamDojo \citep{dreamdojo}. In addition, we construct a stronger baseline, DreamDojo-D, by equipping DreamDojo with a depth decoder and performing full-parameter fine-tuning. As shown in Table \ref{tab:rgb_depth_compare}, our method consistently outperforms all baselines in both RGB and depth prediction. Compared with DreamDojo-D, the results further demonstrate the benefits of multi-view inputs. As illustrated in Fig. \ref{fig: lam}, our method produces more accurate and spatially consistent depth predictions, indicating stronger 3D scene understanding. Extensive analyses in the Appendix B, including action transfer, multi-view attention visualization, mutual information estimation, and linear probing, further validate the effectiveness, robustness, and 3D awareness of the learned latent action representations.

\subsection{Video Quality Evaluation for World Models}

We evaluate the proposed method against SOTA methods on imagined video generation quality using the robotic world model benchmark WorldArena \citep{worldarena}. The compared methods include GigaWorld-0 \citep{team2025gigaworld}, Genie Envisioner \citep{liao2025genie}, RoboMaster \citep{fu2025learning}, WoW \citep{chi2025wow}, IRASim \citep{zhu2025irasim}, Cosmos 2.5 (action) \citep{ali2025world}, and DreamDojo (Post-train). For fair comparison, all baselines with publicly available code are finetuned under the same setting as DreamDojo, using a resolution of (640 $\times$ 480) and 13-frame video sequences. As illustrated in Fig. \ref{fig: arena}, we present qualitative visualization results on two representative sub-tasks. Furthermore, Table \ref{tab:arena} reports quantitative evaluations across six major dimensions of video generation quality. The results demonstrate that our method achieves SOTA performance on the vast majority of evaluation metrics. Although several previous approaches exhibit advantages in visual fidelity, our method remains highly competitive in visual quality while significantly outperforming existing methods in action fidelity. Notably, our method shows particularly strong performance in Motion Quality and 3D Accuracy related metrics. These results indicate that 3D-aware latent action pretraining substantially enhances the world model’s capability in spatial structure understanding and dynamic motion modeling, leading to significant overall performance gains.

\begin{table}[t]
\centering
\caption{Generalization evaluation results on three OOD datasets (Pretraining on human videos).}
\label{tab:ood}

\setlength{\tabcolsep}{3pt}
\renewcommand{\arraystretch}{1.05}

\small
\resizebox{\columnwidth}{!}{
\begin{tabular}{l ccc ccc ccc}
\toprule

\multirow{2}{*}{\textbf{Methods}}
& \multicolumn{3}{c}{\textbf{In-lab}}
& \multicolumn{3}{c}{\textbf{EgoDex}}
& \multicolumn{3}{c}{\textbf{DreamDojo-HV}} \\

\cmidrule(lr){2-4}
\cmidrule(lr){5-7}
\cmidrule(lr){8-10}

& PSNR$\uparrow$
& SSIM$\uparrow$
& LPIPS$\downarrow$

& PSNR$\uparrow$
& SSIM$\uparrow$
& LPIPS$\downarrow$

& PSNR$\uparrow$
& SSIM$\uparrow$
& LPIPS$\downarrow$ \\

\midrule

Cosmos-Predict2.5
& 20.576 & 0.774 & 0.222
& 19.952 & 0.787 & 0.219
& 18.274 & 0.754 & 0.236 \\

\textsc{DreamDojo-2B}
& 21.114 & 0.774 & 0.222
& 20.411 & 0.775 & 0.226
& 18.813 & 0.747 & 0.238 \\

\textsc{DreamDojo-14B}
& 21.413 & 0.788 & 0.208 
& 20.525 & 0.787 & 0.213 
& 18.924 & 0.751 & 0.228 \\

\midrule

\textsc{Ours (2B)}
& \textbf{21.465} & \textbf{0.792} & \textbf{0.207}
& \textbf{20.542} & \textbf{0.796} & \textbf{0.197}
& \textbf{19.034} & \textbf{0.777} & \textbf{0.211} \\

\bottomrule
\end{tabular}
}
\end{table}

\subsection{Generalization Evaluation}

DreamDojo has shown that large-scale human manipulation videos can improve the generalization capability of world models. Building on its OOD evaluation protocol, we further analyze the impact of 3D-aware latent actions on world model generalization. Specifically, we post-train the world model on the GR1\_robot dataset \citep{dreamdojo}, and evaluate rollout performance on three OOD benchmarks: In-lab Eval, EgoDex Eval, and DreamDojo-HV Eval \citep{dreamdojo}. These benchmarks contain unseen scenes, objects, and interaction patterns, providing a challenging testbed for cross-scene generalization.

As shown in Table \ref{tab:ood}, our method consistently outperforms DreamDojo across all OOD benchmarks, demonstrating stronger robustness in unseen environments and complex interactions. Qualitative results further show improved action following and physical consistency (See Fig. \ref{fig: ood} and Fig. A3). We attribute these gains to the proposed 3D-aware latent action modeling, which better captures spatial relationships, object states, and interaction dynamics. As a result, our method generates more plausible action trajectories and more stable interaction outcomes, such as consistent banana handover and stable block placement, indicating stronger 3D spatial understanding beyond superficial motion imitation. Furthermore, Appendix~A provides qualitative analyses of long-horizon world model rollouts as well as an evaluation of their effectiveness for downstream policy learning.

\begin{table}[t]
\centering
\caption{Ablation study on the contribution of different components}
\label{tab:ablation_components}

\renewcommand{\arraystretch}{1.08}
\setlength{\tabcolsep}{2.5pt}

\resizebox{0.95\columnwidth}{!}{
\begin{tabular}{lccc|cc|cc}
\toprule

\multirow{2}{*}{\textbf{Setting}}
& \multirow{2}{*}{\textbf{MV.}}
& \multirow{2}{*}{\textbf{Geo.}}
& \multirow{2}{*}{\textbf{Depth}}
& \multicolumn{2}{c|}{\textbf{LAM}}
& \multicolumn{2}{c}{\textbf{World Model}} \\

\cmidrule(lr){5-6}
\cmidrule(lr){7-8}

&&&
& \textbf{PSNR}$\uparrow$
& \textbf{SSIM}$\uparrow$
& \textbf{PSNR}$\uparrow$
& \textbf{SSIM}$\uparrow$ \\

\midrule

Baseline 
& 
& 
& 
& 29.31 
& 0.886 
& 21.22 
& 0.782 \\

Case1 
& \checkmark
& 
& 
& 28.97 
& 0.872 
& 20.53 
& 0.711 \\

Case2 
& 
& \checkmark
& 
& 29.84 
& 0.906 
& 21.36 
& 0.789 \\

Case3 
& \checkmark
& 
& \checkmark
& 29.69 
& 0.912 
& 21.42 
& 0.788 \\

Case4 
& \checkmark
& \checkmark
& 
& 29.92 
& 0.919 
& 21.64 
& 0.793 \\

Default 
& \checkmark
& \checkmark
& \checkmark
& \textbf{30.56}
& \textbf{0.933}
& \textbf{21.68}
& \textbf{0.797} \\

\bottomrule
\end{tabular}
}

\footnotesize
MV.: Multi-view data.
Geo.: Geometry alignment.
Depth: Depth decoder.

\end{table}

\section{Ablation Studies}
\subsection{Component Analysis} 


We analyze the contribution of each component on the In-Lab dataset, using the LAM in DreamDojo as the baseline. Case1 introduces only multi-view data; Case2 adds geometric representation alignment; Case3 further incorporates depth decoding; Case4 combines multi-view data with geometric alignment; Default denotes our full model.

As shown in Table \ref{tab:ablation_components}, multi-view supervision alone (Case1) does not improve performance, indicating that increased viewpoint diversity introduces additional cross-view discrepancies. In contrast, geometric representation alignment (Case2) consistently improves all metrics, demonstrating its effectiveness in enhancing structural latent action representations. Adding depth decoding (Case3) further improves performance by providing additional geometric constraints. Combining multi-view data with geometric alignment (Case4) achieves stronger results, validating the role of alignment in mitigating view inconsistencies. The full model (Default) achieves the best performance across all metrics, demonstrating that the proposed components are complementary and jointly benefit latent action modeling and world model learning. Additional analyses, including computational complexity and further ablations, are provided in the Appendices B, C, D.

\begin{figure}[t]
\centering
\includegraphics[width=0.95\columnwidth]{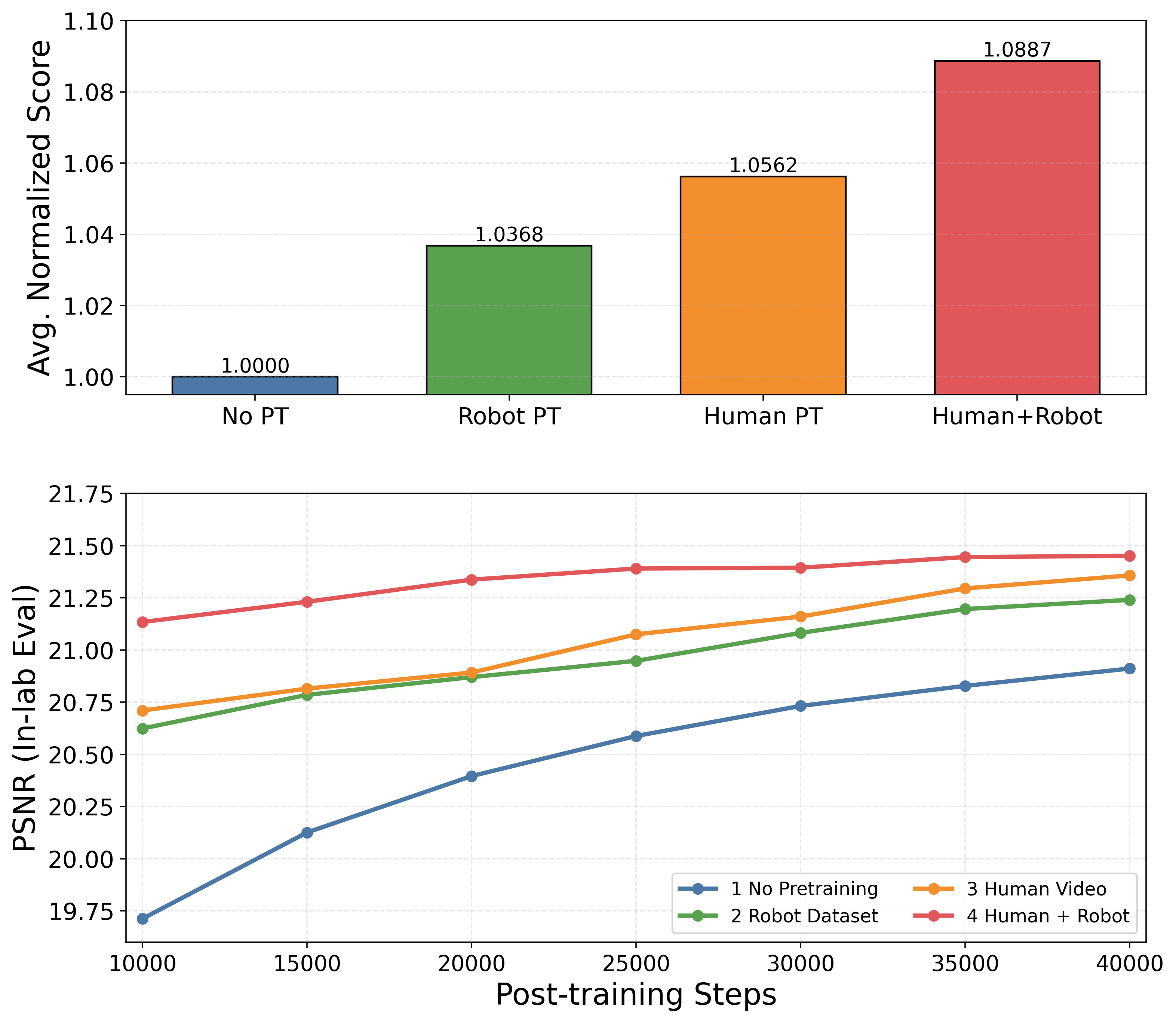}
\caption{Impact of different pretraining strategies on world model performance. (Top) Average normalized scores across evaluation metrics. (Bottom) PSNR convergence curves during post-training.}
\label{fig: data}
\end{figure}


\subsection{Dataset Ablation Analysis} 
We further analyze the impact of different pretraining data sources on downstream adaptation performance of the world model, including: (1) no pretraining; (2) robot-only pretraining; (3) human-video pretraining; and (4) joint human-and-robot video pretraining. As shown in Fig. \ref{fig: data}, all pretrained models outperform the no-pretraining baseline, demonstrating the effectiveness of large-scale video pretraining for world model learning. Notably, human-video pretraining yields larger performance gains than robot-only pretraining, suggesting that diverse human interaction dynamics help learn more generalizable temporal representations. Furthermore, joint human-and-robot pretraining achieves the best overall performance, improving the average normalized score from $1.0000$ to $1.0887$, indicating strong complementarity between the two data sources. In addition, the PSNR convergence curves show that models pretrained on human videos converge substantially faster, suggesting that human-video pretraining provides a better initialization prior, thereby improving downstream prompt fitting efficiency and reducing the number of optimization steps required during adaptation.

\section{Conclusion}

This paper presents LAWM-3D. By leveraging multi-view data, RGB–D co-reconstruction, and geometric feature alignment, LAWM-3D learns view-invariant latent action representations from unlabeled human videos. Experimental results demonstrate that the learned latent actions significantly improve world model training and generalization performance.

\bibliography{aaai2027}

\clearpage
\appendix

\begin{figure}[t]
    \centering
    \includegraphics[width=1\linewidth]{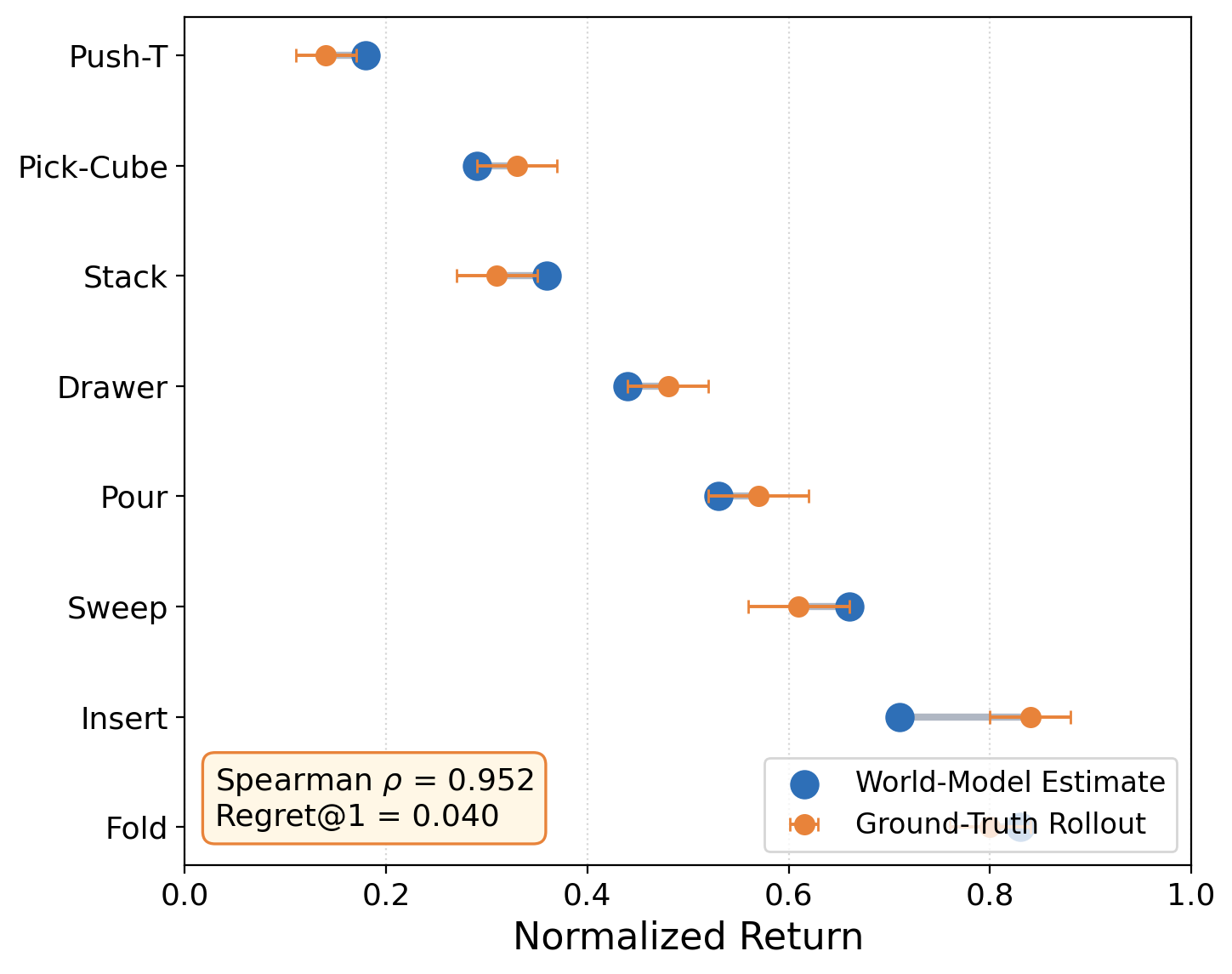}
    \caption{Agreement between imagined returns from the learned world model and real environment performance of candidate policies.}
    \label{fig:policy}
\end{figure}

\begin{figure}[t]
    \centering
    \includegraphics[width=1\linewidth]{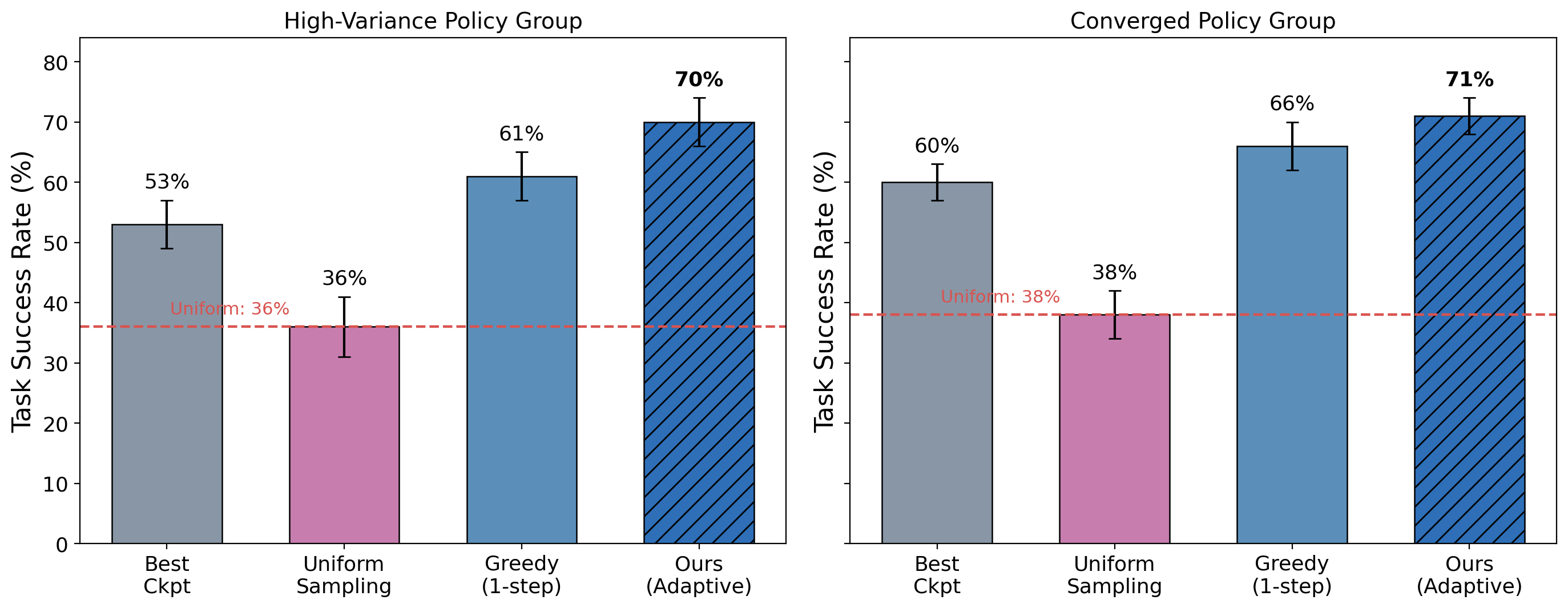}
    \caption{Closed-loop success rates of different policy selection strategies under high-variance and converged policy groups. Our adaptive selection consistently improves performance by exploiting policy diversity.}
    \label{fig:planning}
\end{figure}


In the Appendix, we first present extensive qualitative results on real-world robotic environments to further demonstrate the impact of 3D-aware latent actions on world model performance compared with the baseline approach. We also showcase the applications of the learned world model in policy execution and planning. Second, we conduct comprehensive experiments to analyze and validate the effectiveness of the proposed LAM, including representation analysis, component contribution studies, and ablation studies on key design choices. Third, we provide a complexity analysis to demonstrate that our approach introduces 3D awareness while maintaining efficient training and inference. Finally, we discuss the limitations of our method and potential directions for future improvement.

\begin{figure*}
    \centering
    \includegraphics[width=1\linewidth]{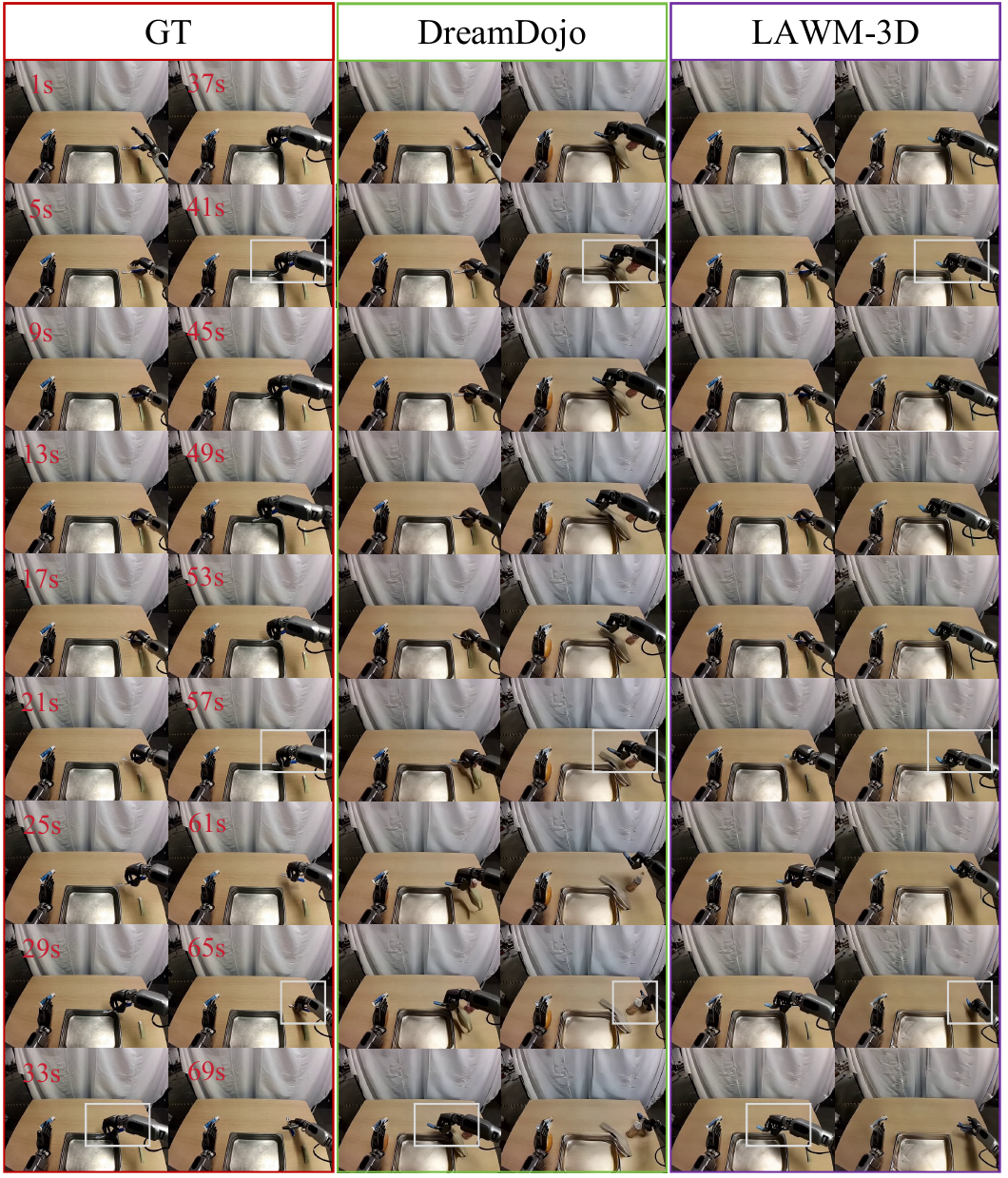}
    \caption{Comparison of long real-world rollouts between ground truth (GT), DreamDojo, and LAWM-3D on manipulation scenarios. In long-horizon rollouts, our generated videos exhibit higher visual quality. Moreover, as highlighted in the annotated regions, LAWM-3D demonstrates better action following, interaction modeling, and 3D understanding. Compared with the baseline, our method achieves stronger stability and consistency, while avoiding non-faithful errors such as visual artifacts.}
    \label{fig:long}
\end{figure*}

\section{A. More Results}

\subsection{Long-Horizon Real-World Rollout Analysis}

To further evaluate the dynamic modeling capability of world models in long-horizon prediction scenarios, we compare the rollout results of our proposed LAWM-3D with the baseline method on real-world robotic manipulation tasks. As shown in the Fig. \ref{fig:long}, LAWM-3D maintains higher visual quality and stronger temporal consistency over long-horizon video generation compared with DreamDojo \citep{dreamdojo}. Specifically, by incorporating 3D-aware latent action representations, our model can more accurately capture the spatial structure of objects, robot motion states, and the interaction relationships between actions and environments, leading to future states that better comply with real-world physical dynamics. As highlighted in the annotated regions, LAWM-3D demonstrates clear advantages in action following, object interaction modeling, and 3D understanding, enabling more stable prediction of critical dynamic changes during robotic manipulation. In contrast, the baseline method tends to accumulate prediction errors over long-horizon rollouts, resulting in non-faithful outcomes such as object state drift, inconsistent interactions, and visual artifacts. These results demonstrate that the 3D-aware latent action space effectively improves the long-term prediction capability and generation stability of world models in complex robotic manipulation scenarios.

\subsection{World Model-based Policy Evaluation and Online Policy Selection}

\paragraph{Experimental Setup.}
All decision-oriented experiments are conducted in our custom-built simulation manipulation suite based on the MuJoCo physics engine \citep{mujoco}. The suite contains eight representative tabletop manipulation tasks, covering pushing, grasping, stacking, articulated object interaction, pouring, sweeping, precision insertion, and deformable object folding, denoted as Push-T, Pick-Cube, Stack, Drawer, Pour, Sweep, Insert, and Fold, respectively.

Each task provides $128 \times 128$ RGB observations from a fixed third-person camera view, together with proprioceptive states. The robot is controlled using a 7-DoF end-effector action space, consisting of Cartesian incremental pose commands and gripper control signals. Each task defines a sparse task-completion reward, and we report the normalized episode return in $[0,1]$, where $1$ indicates successful completion and $0$ denotes failure.

For candidate policies, we train behavior cloning (BC) agents using 200 human teleoperation demonstrations for each task and save checkpoints from different training stages. Based on these checkpoints, we construct two policy sets with different levels of behavioral diversity: a high-variance policy group consisting of under-converged checkpoints, and a converged policy group containing mature checkpoints with similar behaviors.

The world model is trained solely on offline interaction trajectories and remains frozen throughout all evaluations. Therefore, both offline policy evaluation and online policy selection are performed using the same fixed world model without additional environment interaction or model updates.

\paragraph{World Model-based Offline Policy Evaluation.}
We first investigate whether the learned world model can serve as a reliable proxy for evaluating candidate policies without physical execution. For each task, we perform policy rollouts entirely within the learned world model and estimate the expected return from the imagined trajectories. We then compare the predicted returns against the ground-truth performance obtained from real environment rollouts.

As shown in Fig. \ref{fig:policy}, the predicted returns strongly correlate with the actual task performance, achieving a Spearman rank correlation coefficient of $\rho=0.952$. Furthermore, selecting the best checkpoint solely according to the world model prediction results in only $0.040$ Regret@1 in the real environment. These results demonstrate that the learned world model captures decision-relevant dynamics rather than merely generating visually plausible futures, enabling reliable and low-cost policy evaluation.

\paragraph{World Model-based Online Policy Selection.}
Based on the above observation, we further deploy the world model as an online policy selection module. Instead of directly optimizing actions in the high-dimensional action space, we leverage a collection of BC checkpoints as candidate policies, where each checkpoint proposes an action at every control step.

At each decision timestep, the selection module performs multi-step imagination for each candidate action using the learned world model. Specifically, we rollout each proposal for $H=8$ future steps and use a learned value head to score the imagined trajectories. The action proposed by the policy with the highest predicted value is then executed in the environment. In this way, multiple static checkpoints are dynamically combined into a single adaptive controller.

For fair comparison, all baselines share the same frozen world model and identical candidate policy sets, differing only in the strategy used for selecting among proposals:

\begin{itemize}
    \item \textbf{Best Ckpt}: Always executes the single checkpoint with the highest validation success rate.
    \item \textbf{Uniform Sampling}: Randomly samples among candidate policy proposals at each timestep.
    \item \textbf{Greedy (1-step)}: Selects the proposal with the highest immediate predicted reward from one-step imagination.
    \item \textbf{Ours (Adaptive)}: Selects actions based on multi-step imagined returns using the proposed adaptive policy selection mechanism.
\end{itemize}

For each configuration, we conduct 200 closed-loop rollouts and report the average success rate over four random seeds. Error bars indicate standard deviations.

\paragraph{Effect of Policy Diversity.}
To further understand how candidate policy diversity affects the effectiveness of policy selection, we evaluate two different checkpoint groups.

The High-Variance Policy Group consists of checkpoints saved during different training stages, exhibiting diverse behaviors. The Converged Policy Group contains sufficiently trained checkpoints with similar and stable behaviors.

As shown in Fig. \ref{fig:planning}, our adaptive selection strategy achieves the largest improvement on the high-variance policy group. Specifically, it improves the success rate from $53\%$ to $70\%$ compared with the best single checkpoint, yielding a $17\%$ absolute gain and nearly $2\times$ improvement over uniform sampling.

For the converged policy group, although candidate policies are already strong and behaviorally similar, our method still consistently outperforms fixed checkpoint selection and random sampling. These results suggest that the benefit of world model-based policy selection is closely related to the behavioral diversity of candidate policies, and richer policy collections from different architectures or training procedures may further enhance the effectiveness of this framework.

\begin{figure*}[t]
    \centering
    \includegraphics[width=1\linewidth]{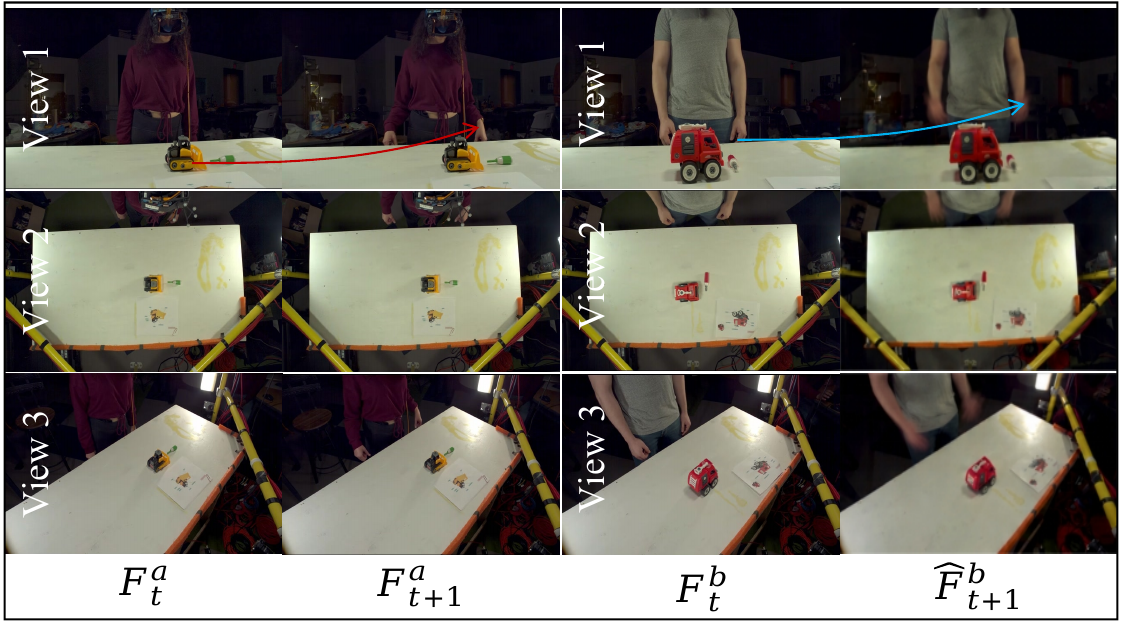}
    \caption{Cross-view transfer of latent actions, showing that the learned action representation captures view-independent motion dynamics.}
    \label{fig:ac}
\end{figure*}

\begin{figure}
    \centering
    \includegraphics[width=1\linewidth]{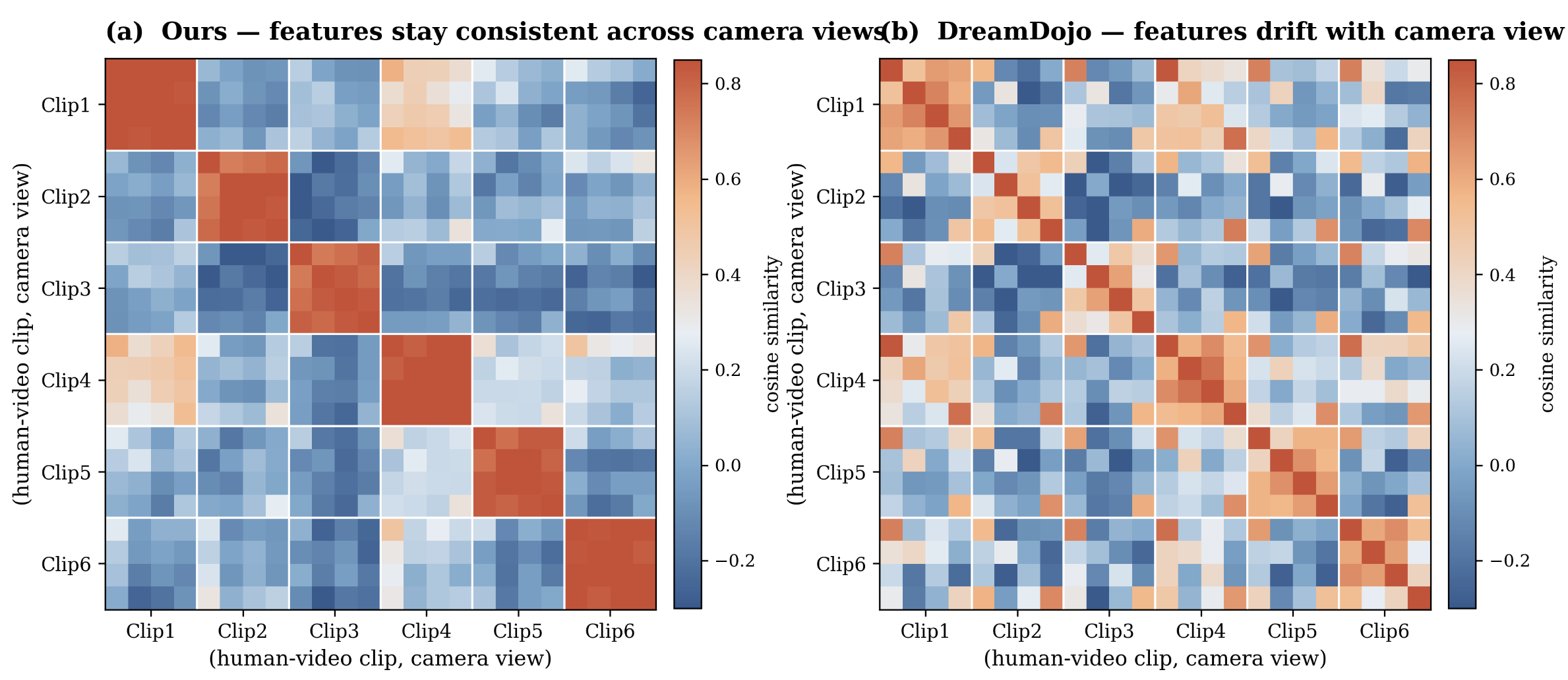}
    \caption{Cosine similarity matrix of latent representations across 100 human video clips. Our method achieves stronger viewpoint invariance while preserving inter-clip discriminability.
}
    \label{fig:cross}
\end{figure}

\section{B. Depth Analysis of Our Latent Action Model}

\subsection{Effectiveness Analysis of Multi-View Latent Actions}

We investigate whether multi-view supervision enables the latent action representation to capture view-invariant motion dynamics rather than viewpoint-specific appearance changes. We conduct two complementary analyses.

First, we evaluate the cross-view transferability of the learned latent actions. Specifically, given two consecutive observations from one viewpoint, we extract the latent action $a_t$ from $F^a_t$ and $F^a_{t+1}$, and combine it with the observation feature from another viewpoint $F^b_t$ to predict the future feature $\hat{F}^b_{t+1}$. As shown in Fig. \ref{fig:ac}, the predicted future representation preserves the motion dynamics of the target viewpoint, demonstrating that the learned latent action encodes transferable temporal information shared across different views.

Second, we analyze the viewpoint invariance of the learned latent action space. We randomly sample 100 human video clips and extract a 32-dimensional representation for each clip-view pair. We then compute pairwise cosine similarities among all representations. As illustrated in Fig. \ref{fig:cross}, our method exhibits a clearer block-diagonal similarity structure, indicating that representations from the same underlying motion remain closer across viewpoints, while different clips remain distinguishable. This suggests that multi-view training effectively improves viewpoint robustness while preserving action discriminability.

\begin{figure}[t]
    \centering
    \includegraphics[width=1\linewidth]{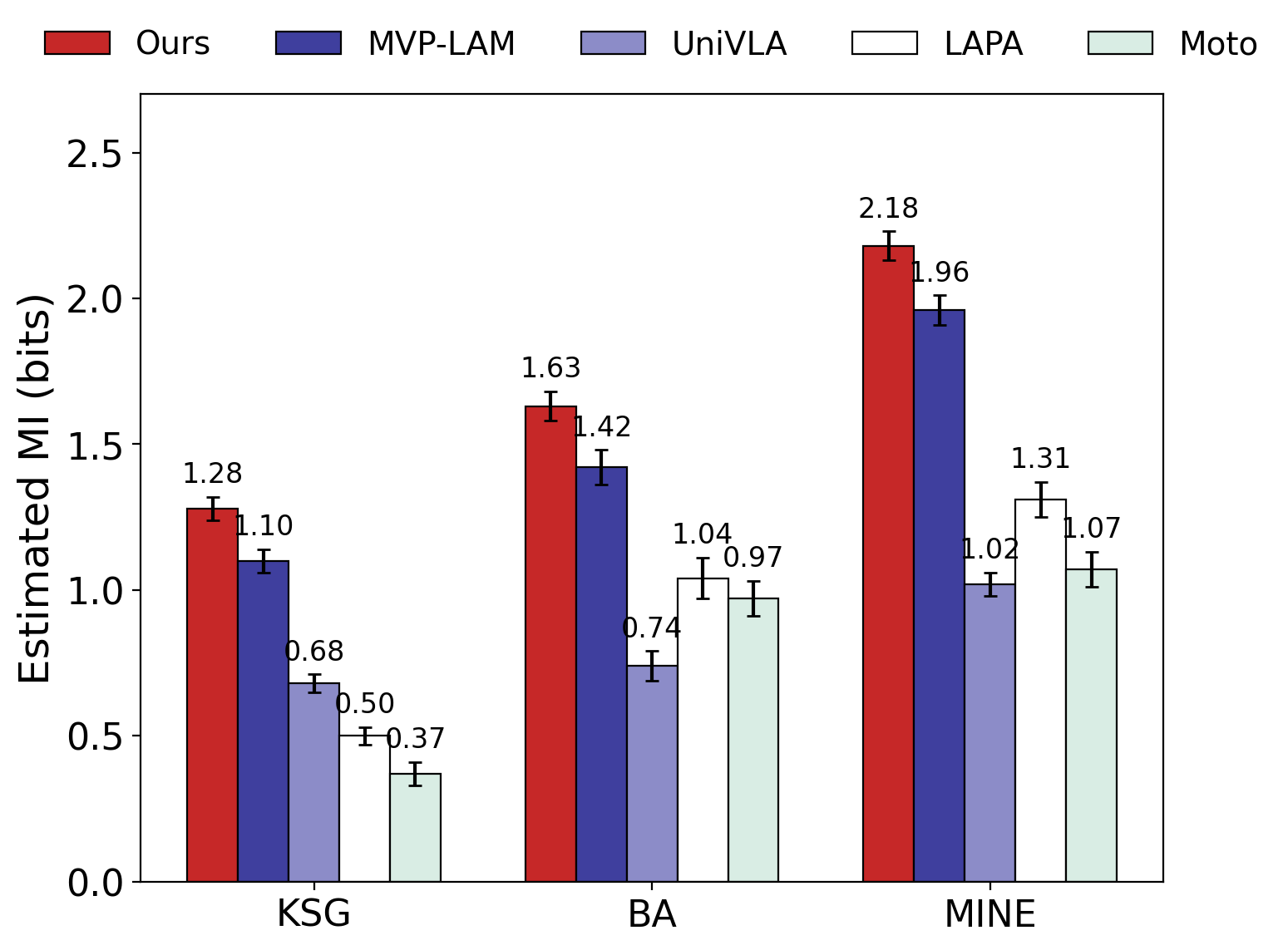}
    \caption{Mutual information analysis of learned latent actions. Our 3D-aware latent action representation achieves higher mutual information with ground-truth actions across different estimators, indicating richer action-related information}
    \label{fig:mu}
\end{figure}

\begin{figure}[t]
    \centering
    \includegraphics[width=1\linewidth]{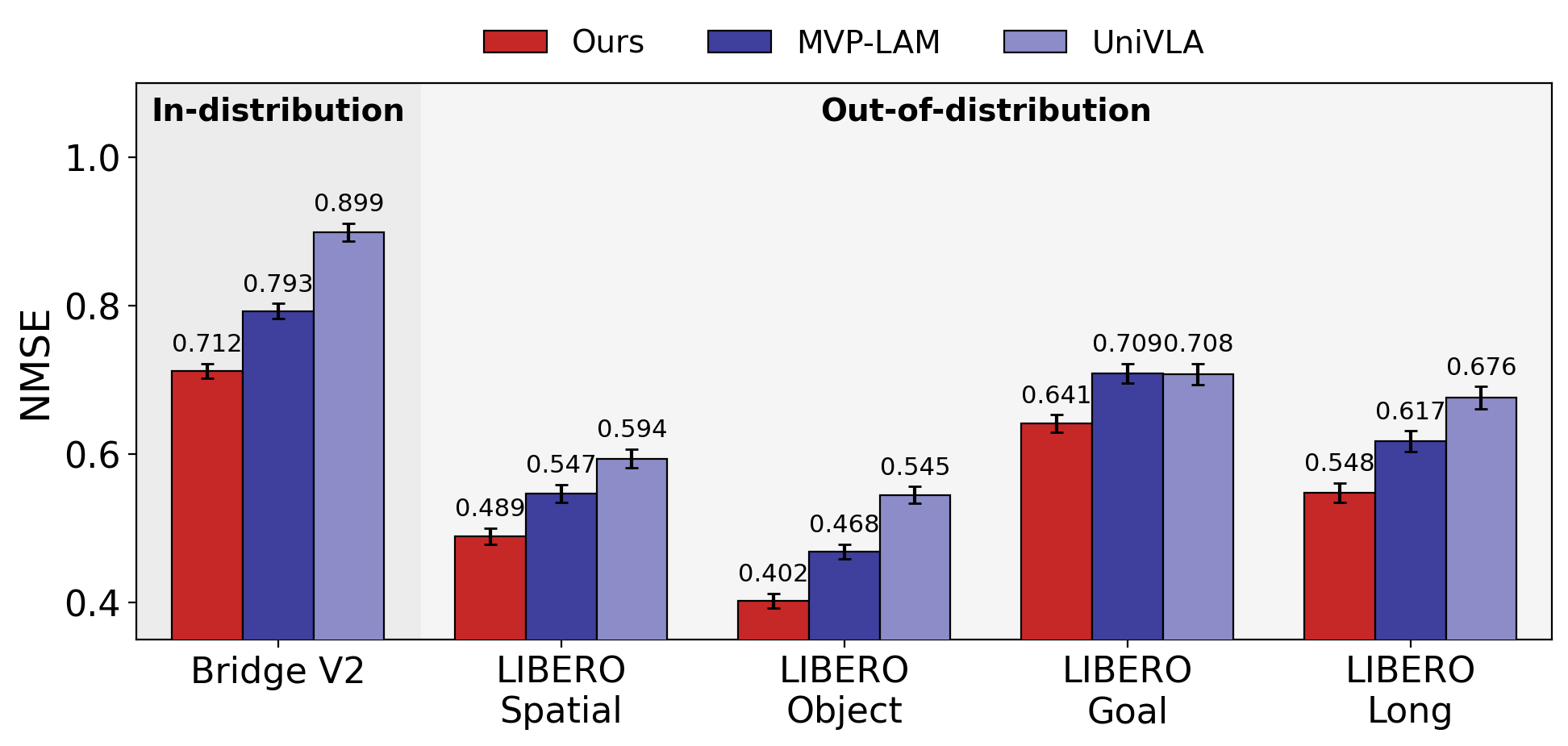}
    \caption{Linear probing evaluation of latent action representations. Our method achieves lower action reconstruction error on both in-distribution and out-of-distribution benchmarks, demonstrating more transferable 3D-aware action representations.}
    \label{fig:li}
\end{figure}

\subsection{3D-Aware Latent Action Space Analysis}


To verify whether the proposed 3D-aware latent action modeling mechanism can learn a more structured and action-relevant latent space, we conduct two complementary analyses: Action Information Quantification and Action Representation Evaluation. Unlike conventional latent action learning approaches that rely solely on single-view video reconstruction, our method introduces cross-view reconstruction constraints and geometric representation alignment, encouraging the latent actions to capture not only observable changes but also viewpoint-invariant dynamic factors associated with underlying physical motions. Therefore, following the evaluation protocol of MVP-LAM \citep{lee2026mvp}, we further investigate whether the proposed 3D-aware mechanism can induce a more action-centric and physically grounded latent action space.

\paragraph{Action Information Quantification.} We first quantify the amount of action-related information contained in the learned latent action space by estimating the mutual information between the latent action representations and the ground-truth robot actions. Specifically, we compute the mutual information between the model-generated latent actions $Z$ and the real robot actions $A$:
\begin{equation}
I(Z;A).
\end{equation}
A higher mutual information value indicates that the latent representation contains richer action-relevant information, suggesting a stronger capability to capture dynamic changes induced by physical interactions. This metric directly measures whether the learned latent action space focuses on underlying physical motions rather than merely encoding viewpoint variations or appearance-related factors.

We conduct experiments on the Bridge V2 dataset \citep{walke2023bridgedata}. Since mutual information between high-dimensional continuous variables is difficult to estimate directly, we employ three mutual information estimators with different theoretical foundations to ensure the reliability of our analysis. Specifically, Kraskov--St\"ogbauer--Grassberger (KSG) adopts a non-parametric estimation strategy based on k-nearest neighbors. Considering its estimation instability in high-dimensional spaces, we first reduce the latent action representations to 256 dimensions using random projection before applying KSG estimation. The Barber--Agakov (BA) estimator learns a conditional distribution $q(a|z)$ to construct a variational lower bound of mutual information, while MINE~\citep{belghazi2018mutual} estimates mutual information based on the Donsker--Varadhan representation using a neural network.

As shown in Fig. \ref{fig:mu}, our method consistently achieves the highest mutual information scores across all three estimators, outperforming the strongest baseline MVP-LAM by an average margin of approximately 15\%. This consistent improvement demonstrates that, by incorporating cross-view constraints and 3D geometric supervision, our model learns a more compact latent representation that is highly correlated with real physical actions. In contrast to conventional approaches that may entangle viewpoint changes and background variations with action information, the proposed 3D-aware latent space more effectively captures viewpoint-consistent dynamic factors underlying physical motions.

\paragraph{Action Representation Evaluation.} To further verify whether the learned latent action representations preserve structured information that is useful for action prediction, we adopt a linear probing protocol to evaluate the action recoverability of different latent spaces.

Specifically, after freezing the learned latent action representation $z_t$, we only train a linear mapping:
\begin{equation}
\hat{a}_t = W z_t + b,
\end{equation}
to predict the corresponding ground-truth action $a_t$. We use the Normalized Mean Squared Error (NMSE) as the evaluation metric, where a lower error indicates that the action information can be more accurately recovered through a simple linear transformation, suggesting that the latent space preserves more explicit and structured action-related information.

Since different methods produce latent representations with varying dimensions, we first apply PCA to project all latent features into a unified 128-dimensional space before performing linear probing to ensure a fair comparison. Furthermore, following the evaluation protocol of MVP-LAM \citep{lee2026mvp}, we eliminate the influence of different action horizons among methods by converting all predicted actions into net relative actions. Specifically, we first denormalize the actions using the statistics of the original dataset, then accumulate the actions according to the corresponding prediction horizon $H$ of each method, and finally re-normalize them using the statistics of the target horizon. This procedure ensures that all methods are evaluated under the same physical action prediction objective.

Experiments are conducted on both the in-distribution Bridge V2 dataset and several out-of-distribution (OOD) tasks \citep{liu2023libero}, including LIBERO-Spatial, LIBERO-Object, LIBERO-Goal, and LIBERO-Long, to evaluate the generalization capability of the learned 3D-aware latent action space.

As shown in Fig. \ref{fig:li}, our method achieves the lowest NMSE on Bridge V2 and consistently outperforms all baseline methods across the OOD tasks. In particular, the performance advantage becomes more significant on the LIBERO-Long task, which involves longer temporal horizons and more complex dynamic variations. These results demonstrate that the latent action representations learned through cross-view consistency constraints and 3D geometric supervision not only encode action semantics more accurately but also capture physically meaningful dynamic structures that generalize across environments, leading to a more robust and physically grounded 3D-aware action space.

\begin{table}[t]
\centering
\caption{Effect of different latent action aggregation strategies.}
\label{tab:aggregation}
\resizebox{0.48\textwidth}{!}{
\begin{tabular}{lcccc}
\toprule
Aggregation & RGB PSNR$\uparrow$ & RGB SSIM$\uparrow$ & Depth PSNR$\uparrow$ & Depth SSIM$\uparrow$ \\
\midrule
Max Pooling       & 33.08 & 0.924 & 35.84 & 0.931 \\
Attention Pooling & 33.34 & 0.929 & 36.21 & 0.939 \\
Concatenation     & 33.29 & 0.928 & 36.08 & 0.937 \\
Mean Pooling      & \textbf{33.56} & \textbf{0.933} & \textbf{36.58} & \textbf{0.945} \\
\bottomrule
\end{tabular}
}
\end{table}

\subsection{Analysis of Latent Action Aggregation}

LAWM-3D aggregates action tokens extracted from multiple viewpoints into a unified latent action representation. To analyze the impact of different aggregation strategies, we compare four methods: Mean Pooling, Max Pooling, Attention Pooling, and Concatenation. As shown in Table~\ref{tab:aggregation}, Mean Pooling achieves superior performance on both RGB and depth reconstruction tasks. In contrast, Attention Pooling tends to bias towards a few dominant viewpoints, thereby compromising cross-view consistency; Concatenation preserves independent representations for each viewpoint, introducing feature redundancy that hinders the learning of a unified action space; and Max Pooling retains only the strongest local responses, often resulting in the loss of continuous motion dynamics and fine-grained geometric details. These results indicate that Mean Pooling is the most suitable strategy for unifying multi-view action information, facilitating the learning of more robust 3D perceptual latent action representations. We attribute this advantage to the fact that while different viewpoints exhibit distinct visual appearances, they correspond to the same underlying action; thus, their action representations should share consistent semantics. Mean Pooling effectively aggregates the dynamic information shared across viewpoints while mitigating noise introduced by viewpoint variations, occlusions, and lighting differences, ultimately yielding a more stable and view-consistent latent action representation.

\section{C. Depth Analysis of Multi-View}

\begin{figure*}
    \centering
    \includegraphics[width=1\linewidth]{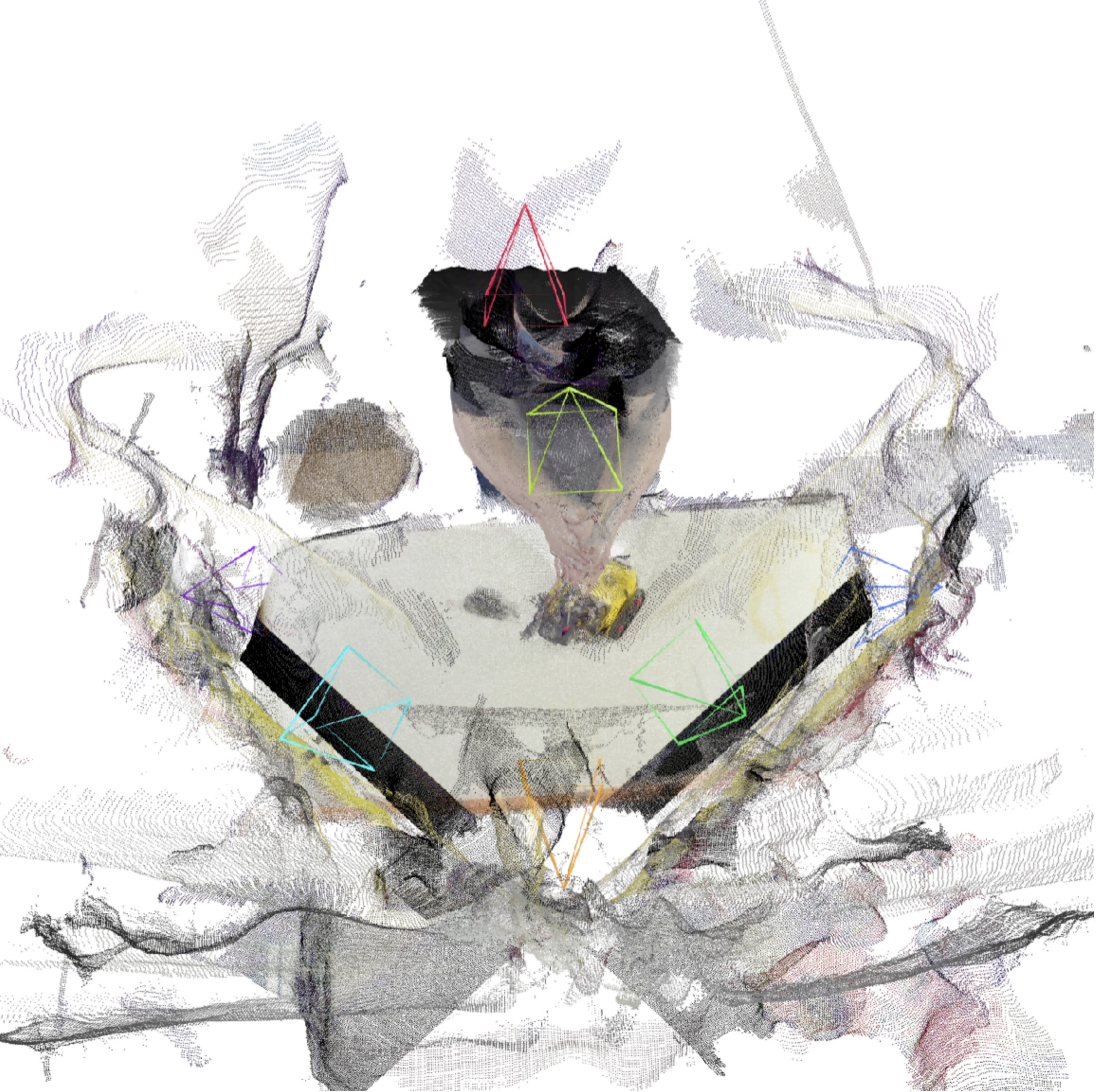}
    \caption{Visualization of VGGT point cloud reconstruction of synchronized multi-view human manipulation videos.}
    \label{fig:vggt}
\end{figure*}

\subsection{Visualization of VGGT Result}

To validate whether the pre-trained VGGT can provide reliable 3D geometric supervision for latent action learning, we first evaluate its capability in geometric reconstruction of multi-view videos. Specifically, we input synchronized multi-view videos from the Assembly101 dataset into a VGGT model with frozen parameters, without any fine-tuning. As illustrated in Fig. \ref{fig:vggt}, VGGT accurately reconstructs the 3D scene structure across diverse viewpoints. These results strongly substantiate the feasibility of leveraging multi-view human videos for VGGT alignment.

\subsection{Analysis of Geometric Alignment}

We further conduct a set of controlled experiments to answer three key questions regarding the design of the geometric representation alignment: (i) which encoder layers should be aligned with VGGT, (ii) how sensitive the model is to the alignment loss weights, and (iii) whether VGGT is indispensable or other 3D foundation models can achieve comparable effects. All experiments below are conducted on the In-Lab dataset, and we report the LAM reconstruction quality (RGB PSNR/SSIM) as the primary metric.

\begin{table}[t]
\centering
\caption{Effect of aligning different ranges of encoder layers with VGGT. $h=[a,b]$ denotes aligning all consecutive layers from the $a$-th to the $b$-th, with layers indexed from $1$ to $L=24$.}
\label{tab:layer}
\resizebox{0.46\textwidth}{!}{
\begin{tabular}{lcc}
\toprule
Aligned Layers $h$ & RGB PSNR$\uparrow$ & RGB SSIM$\uparrow$ \\
\midrule
None (Rec. only)      & 29.31 & 0.886 \\
$[1,6]$ (shallow)     & 29.74 & 0.905 \\
$[19,24]$ (deep)      & 29.96 & 0.917 \\
$[6,8]$ (narrow)      & 30.15 & 0.922 \\
$[6,11]$ (ours)       & \textbf{30.56} & \textbf{0.933} \\
$[6,16]$ (wide)       & 30.38 & 0.927 \\
$[1,24]$ (all layers) & 30.02 & 0.918 \\
\bottomrule
\end{tabular}
}
\end{table}

\paragraph{Which layers to align?}
The encoder consists of $L=24$ transformer layers, and we align all consecutive layers within a range $h=[a,b]$ with the corresponding VGGT features. We vary this range and report the results in Table~\ref{tab:layer}. Aligning only the shallow range ($[1,6]$) yields limited improvement, as early features mainly encode low-level appearance cues that are weakly correlated with 3D geometry. Aligning only the deep range ($[19,24]$) is also suboptimal, since deep features become increasingly task-specific and abstract, and forcing them to match VGGT geometry over-constrains the action-relevant representation. Aligning the intermediate range $h=[6,11]$ achieves the best trade-off, because these layers already capture spatial-semantic structures while retaining sufficient flexibility to encode dynamic action information. A too narrow range ($[6,8]$) provides insufficient geometric supervision, whereas an overly wide range ($[6,16]$ or the full $[1,24]$) forces excessive geometric alignment and reduces the capacity available for action modeling, slightly degrading performance. These results justify our choice of the intermediate range $h=[6,11]$ as a balanced supervision region.

\begin{table}[t]
\centering
\caption{Sensitivity to alignment loss weights ($\lambda_{\text{Angular}}$, $\lambda_{\text{Scale}}$).}
\label{tab:weight}
\resizebox{0.46\textwidth}{!}{
\begin{tabular}{cccc}
\toprule
$\lambda_{\text{Angular}}$ & $\lambda_{\text{Scale}}$ & RGB PSNR$\uparrow$ & RGB SSIM$\uparrow$ \\
\midrule
0.0  & 0.0   & 29.31 & 0.886 \\
0.1  & 0.05  & 30.18 & 0.921 \\
0.5  & 0.05  & \textbf{30.56} & \textbf{0.933} \\
1.0  & 0.05  & 30.37 & 0.926 \\
0.5  & 0.01  & 30.29 & 0.924 \\
0.5  & 0.1   & 30.33 & 0.927 \\
0.5  & 0.5   & 29.95 & 0.915 \\
\bottomrule
\end{tabular}
}
\end{table}

\paragraph{Are the alignment weights sensitive?}
Table~\ref{tab:weight} reports the effect of varying the angular and scale loss weights. Overall, the model is robust to a wide range of weights and consistently outperforms the reconstruction-only baseline. The angular loss dominates the geometric alignment: a moderate value ($\lambda_{\text{Angular}}=0.5$) provides sufficient directional supervision, whereas an overly large value ($1.0$) begins to compete with the reconstruction objective and slightly reduces quality. The scale loss plays an auxiliary role in recovering magnitude information; a small weight ($\lambda_{\text{Scale}}=0.05$) is beneficial, while a large value ($0.5$) makes the training unstable and degrades performance, consistent with the observation that directly regressing full-magnitude geometric features is difficult. These results confirm that our default setting ($\lambda_{\text{Angular}}=0.5$, $\lambda_{\text{Scale}}=0.05$) is both effective and stable, and that the method does not require careful per-dataset tuning.

\begin{table}[t]
\centering
\caption{Comparison of different 3D foundation models used as the alignment target.}
\label{tab:foundation}
\resizebox{0.46\textwidth}{!}{
\begin{tabular}{lcc}
\toprule
Alignment Target & RGB PSNR$\uparrow$ & RGB SSIM$\uparrow$ \\
\midrule
None (Rec. only)          & 29.31 & 0.886 \\
DINOv2 (2D semantic)      & 29.78 & 0.907 \\
Depth-Anything-V2         & 30.02 & 0.916 \\
DUSt3R                    & 30.31 & 0.926 \\
VGGT (ours)               & \textbf{30.56} & \textbf{0.933} \\
\bottomrule
\end{tabular}
}
\end{table}

\paragraph{Is VGGT indispensable?}
To examine whether the improvement is specific to VGGT, we replace the alignment target with several alternative pre-trained models, including a 2D semantic model (DINOv2 \cite{oquab2023dinov2}), a monocular depth estimator (Depth-Anything-V2 \cite{yang2024depth}), and a multi-view geometry model (DUSt3R \cite{wang2024dust3r}). As shown in Table~\ref{tab:foundation}, all geometry-aware targets improve over the reconstruction-only baseline, demonstrating that the benefit stems from injecting 3D geometric priors rather than from a particular model. Aligning with the purely 2D semantic features of DINOv2 yields the smallest gain, as it lacks explicit cross-view geometric consistency. Multi-view geometry models (DUSt3R and VGGT) achieve the largest improvements, since their features inherently encode view-consistent 3D structures. VGGT performs best because its alternating intra-frame and inter-frame attention jointly models geometry and multi-view correspondence, providing the most complete geometric supervision. This confirms that VGGT is the most effective choice, while our framework remains compatible with other 3D foundation models.

\subsection{Multi-View Gain or Alignment Gain?}

A natural question is whether the performance gains of LAWM-3D primarily originate from the additional multi-view data or from the geometric feature alignment. As reported in the component ablation of the main paper (Table 4), naively introducing multi-view data alone (Case1) does \textit{not} improve performance and even slightly degrades it, because different viewpoints introduce cross-view appearance discrepancies that the reconstruction objective cannot resolve. In contrast, applying geometric alignment alone (Case2) consistently improves all metrics even under the single-view setting, indicating that the alignment mechanism itself is the primary source of the improvement.

More importantly, the two factors are complementary rather than independent. When multi-view data is combined with geometric alignment (Case4), the model substantially outperforms both single-factor variants, and the full model with depth decoding (Default) achieves the best results. This reveals a clear synergy: geometric alignment provides the cross-view consistency needed to exploit multi-view data effectively, while multi-view data in turn supplies richer geometric evidence that strengthens the alignment. Therefore, the improvement of LAWM-3D is not merely a byproduct of more data, but rather stems from the geometric alignment that transforms multi-view redundancy into view-invariant, physically grounded latent action representations.

\begin{table*}[t]
\centering
\caption{
\textbf{Computational cost analysis of LAWM-3D.}
The proposed framework introduces two additional frozen modules, i.e., LAM ($\sim$710M) and Depth-Anything-V2 ($\sim$97M), during world-model training.
Only the LAM encoder is used to extract latent actions, while the depth model is used only for offline supervision.
Neither module is involved during rollout inference.
}
\label{tab:cost}
\resizebox{\textwidth}{!}{
\begin{tabular}{l c c c c c}
\toprule
\textbf{Component} &
\textbf{Params} &
\textbf{Trainable?} &
\textbf{Training Usage} &
\textbf{Rollout} &
\textbf{Extra VRAM} \\
&
&
&
&
&
\textbf{(FP16)} \\
\midrule
\multicolumn{6}{l}{\textit{World model components}}\\
\midrule
DiT (2B) &
2B &
\checkmark &
Backbone &
\checkmark &
$\sim$4 GB \\

DiT (14B) &
14B &
\checkmark &
Backbone &
\checkmark &
$\sim$28 GB \\

Wan2.1 VAE Tokenizer &
270M &
$\times$ &
Encode/decode latent &
\checkmark &
$\sim$0.5 GB \\

UMT5-XXL Text Encoder &
4.7B &
$\times$ &
Text embedding &
\checkmark &
$\sim$9 GB \\

\midrule
\multicolumn{6}{l}{\textit{Additional modules}}\\
\midrule

LAM Encoder (24-layer ST-Attn) &
405M &
$\times$ &
1 forward / step &
$\times$ &
$\sim$0.8 GB \\

LAM Decoder (24-layer S-Attn) &
304M &
$\times$ &
Pre-training only$^{\dagger}$ &
$\times$ &
$\sim$0.6 GB \\

LAM FC + Projection &
0.9M &
$\times$ &
1 forward / step &
$\times$ &
$<$1 MB \\

Depth-Anything-V2-Base &
97M &
$\times$ &
Offline preprocessing$^{\ddagger}$ &
$\times$ &
$\sim$0.2 GB \\

\midrule
\midrule

\textbf{Configuration}
&
\textbf{Trainable}
&
\textbf{Frozen}
&
\textbf{Training}
&
\textbf{Rollout}
&
\textbf{Peak VRAM}
\\

&
\textbf{Params}
&
\textbf{Params}
&
\textbf{Speed}
&
\textbf{Latency}
&
\textbf{(2B, 1 GPU)}
\\

\midrule

Single-view (2B)
&
2B
&
5.77B
&
$\sim$10.4 s/iter$^*$
&
1.0$\times$
&
$\sim$55 GB
\\

Multi-view (7-view, 2B)
&
2B
&
5.77B
&
$\sim$10.4 s/iter$^*$
&
1.0$\times^{**}$
&
$\sim$62 GB
\\

Single-view (14B)
&
14B
&
5.77B
&
---
&
1.0$\times$
&
---
\\

\bottomrule
\end{tabular}
}
\vspace{2mm}

\footnotesize
$^{\dagger}$ The LAM decoder is only used during LAM pre-training for frame reconstruction and is not involved in world-model training.

$^{\ddagger}$ Depth maps are pre-computed offline and the depth estimator is not loaded during training or rollout inference.

$^{*}$ The additional LAM encoder forward pass contributes less than 0.1 s/iteration, corresponding to less than 1\% of the total training time.

$^{**}$ During multi-view rollout, LAM is not invoked. The only additional cost comes from longer latent sequences, which are handled by context-parallel DiT.
\end{table*}

\section{D. Computational Cost Analysis}
Table~\ref{tab:cost} summarizes the computational overhead of LAWM-3D during both training and rollout inference. In addition to the world model, LAWM-3D introduces two auxiliary modules: a pre-trained latent action model (LAM, approximately 710M parameters) and Depth-Anything-V2 (approximately 97M parameters). Both modules remain frozen throughout world-model training and do not introduce additional trainable parameters. Specifically, only the LAM encoder is used to extract latent action representations during training, while the decoder is used exclusively during LAM pre-training. Depth-Anything-V2 is employed only to generate offline depth supervision and is therefore not loaded during either world-model training or rollout inference.

The additional computation during training is limited to a single forward pass through the frozen LAM encoder, without gradient computation or parameter updates. Consequently, the computational overhead is negligible, contributing less than 1\% to the overall training time per iteration. Under the multi-view setting, the primary computational increase arises from the longer latent token sequences introduced by multiple camera views rather than from additional network computation. This overhead is efficiently handled by the context-parallel implementation of the DiT backbone, resulting in good scalability as the number of views increases. The additional GPU memory consumption mainly comes from storing the frozen module parameters and their forward activations, corresponding to approximately 2 GB in the single-view setting and approximately 9 GB in the seven-view setting.

Importantly, the additional modules introduced by LAWM-3D are not involved during rollout inference. The trained world model performs autoregressive prediction directly using latent actions provided by the policy, without invoking the LAM encoder, while Depth-Anything-V2 is likewise absent from the inference pipeline. As a result, LAWM-3D introduces no additional rollout latency. Furthermore, LAM pre-training is a one-time offline procedure. Once trained, the LAM is frozen and can be reused across different downstream world models and tasks without retraining. Therefore, the additional computational cost of LAWM-3D is confined to a small amount of training-time forward computation and GPU memory overhead, while leaving the deployment efficiency and rollout inference speed unchanged.

\section{Limitations}

While LAWM-3D demonstrates strong performance in learning view-invariant and physically grounded latent action representations, several limitations remain and point to promising directions for future work.

\paragraph{Dependence on pre-trained 3D foundation models.}
Our geometric alignment relies on a frozen 3D foundation model (VGGT) to provide geometric supervision. Consequently, the quality of the learned representations is upper-bounded by the geometric priors encoded in the foundation model. In scenes that fall outside the training distribution of VGGT, such as heavy occlusions, transparent or reflective objects, and extreme viewpoints, the geometric features may become unreliable and provide weaker supervision. Although our experiments show that the framework is compatible with alternative 3D foundation models, a systematic study of how foundation-model errors propagate into the latent action space is left for future work.

\paragraph{Reliance on synchronized multi-view data.}
The multi-view supervision and cross-view consistency constraints require temporally synchronized recordings from multiple calibrated viewpoints. Such multi-view human manipulation data (e.g., Assembly101) is still relatively scarce compared to the vast amount of single-view internet videos. This dependence limits the scale of pre-training data and may restrict the diversity of scenes and actions the model can observe. Extending the alignment mechanism to leverage unsynchronized or single-view videos, for instance through learned view synthesis or pseudo-multi-view generation, is an important direction.

\paragraph{Offline depth supervision.}
The depth signals used for the co-reconstruction objective are pre-computed offline by a monocular depth estimator (Depth-Anything-V2). These predicted depth maps are inherently noisy and only provide relative rather than metric depth, which may introduce scale ambiguity and inconsistencies across frames and views. Incorporating metric or multi-view-consistent depth, or jointly refining depth during training, could further strengthen the geometric constraints.

\paragraph{Training cost and model scale.}
Although the auxiliary modules introduce negligible overhead during rollout inference, pre-training the LAM and the world model from scratch still requires substantial computational resources (e.g., large batch sizes and hundreds of thousands of iterations on high-end GPUs). Due to resource constraints, our largest world model experiments are limited to the 2B backbone, and the scaling behavior of the proposed 3D-aware modeling at larger scales (e.g., 14B) remains to be fully explored.

\paragraph{Scope of evaluation.}
Our evaluation focuses on human and robotic manipulation scenarios in laboratory and OOD benchmarks. The generalization of the learned 3D-aware latent action space to more diverse embodiments, dynamic and deformable objects, and long-horizon compositional tasks has not been extensively validated. We regard closing this gap, together with real-world closed-loop deployment, as an important avenue for future research.


\end{document}